\documentclass{article}

\usepackage{microtype}
\usepackage{graphicx}
\usepackage{subfigure}
\usepackage{booktabs} 
\usepackage{bigstrut}
\usepackage{mathtools}
\usepackage{amsmath,amsfonts}
\usepackage[ruled]{algorithm2e}
\usepackage{bigstrut}
\usepackage{booktabs}
\usepackage[font=small]{caption}
\usepackage{multirow}
\usepackage{tabulary}
\usepackage{balance}

\usepackage{hyperref}

\newcommand{\methodname}{\textsc{Pufferfish}}
\newcommand{\eg}{{\it e.g.}, }
\newcommand{\ie}{{\it i.e.}, }

\usepackage[accepted]{mlsys2021}

\mlsystitlerunning{\textsc{Pufferfish}: Communication-efficient Models At No Extra Cost}

\begin{document}

\twocolumn[
\mlsystitle{\textsc{Pufferfish}: Communication-efficient Models At No Extra Cost}




\begin{mlsysauthorlist}
\mlsysauthor{Hongyi Wang,}{uwcs}
\mlsysauthor{Saurabh Agarwal,}{uwcs}
\mlsysauthor{Dimitris Papailiopoulos}{uwece}
\end{mlsysauthorlist}

\mlsysaffiliation{uwcs}{Department of Computer Sciences, University of Wisconsin-Madison,}
\mlsysaffiliation{uwece}{Department of Electrical and Computer Engineering, University of Wisconsin-Madison}

\mlsyscorrespondingauthor{Hongyi Wang}{hongyiwang@cs.wisc.edu}

\mlsyskeywords{Machine Learning, MLSys}

\vskip 0.3in

\begin{abstract}
    To mitigate communication overheads in distributed model training, several studies propose
    the use of compressed stochastic gradients, usually achieved by sparsification or quantization. 
    Such techniques achieve high compression ratios, but in many cases incur either significant computational overheads or some accuracy loss.
    In this work, we present \methodname{}, a communication and computation efficient distributed training framework that incorporates the gradient compression into the model training process via training low-rank, pre-factorized deep networks. 
    \methodname{} not only reduces communication, but also completely bypasses any computation overheads related to compression, and achieves the same accuracy as state-of-the-art, off-the-shelf deep models. 
    \methodname{} can be directly integrated into current deep learning frameworks with minimum implementation modification. 
    Our extensive experiments over real distributed setups, across a variety of large-scale machine learning tasks, indicate that \methodname{} achieves up to $1.64\times$ end-to-end speedup over the latest distributed training API in PyTorch without accuracy loss.   Compared to the \textit{Lottery Ticket Hypothesis} models, \methodname{} leads to equally accurate, small-parameter models while avoiding the burden of ``winning the lottery''. 
    \methodname{} also leads to more accurate and smaller models than  SOTA structured model pruning methods. 
\end{abstract}
]



\printAffiliationsAndNotice{} 

\section{Introduction}\label{sec:intro}
	Distributed model training plays a key role in the success of modern machine learning systems. Data parallel training, a popular variant of distributed training, has demonstrated massive speedups in real-world machine learning applications and systems \citep{li2014scaling,dean2012large,chen2016revisiting}. Several machine learning frameworks such as TensorFlow \citep{abadi2016tensorflow} and PyTorch \cite{paszke2019pytorch} come with distributed implementations of popular training algorithms, such as mini-batch SGD. However, the empirical speed-ups offered by distributed training, often fall short of a best-case linear scaling. It is now widely acknowledged that communication overheads are one of the key sources of this saturation phenomenon~\cite{dean2012large, seide20141, strom2015scalable, qi17paleo,grubic2018synchronous}.
	 
    Communication bottlenecks are attributed to frequent gradient updates, transmitted across compute nodes. As the number of parameters in state-of-the-art (SOTA) deep models scales to hundreds of billions, the size of communicated gradients scales proportionally~\cite{he2016deep, huang2017densely,devlin2018bert,devlin2019bert,brown2020language}.
	 To reduce the cost of communicating model updates, recent studies propose compressed versions of the computed gradients. A large number of recent studies revisited the idea of low-precision training as a means to reduce communication~\cite{
	 seide20141,
	 	de2015taming,
	 	alistarh2017qsgd,
	 	zhou2016dorefa,
	 	wen2017terngrad,
	 	zhang2017zipml,
	 	de2017understanding,
	 	de2018high,
	 	bernstein2018signsgd,
	 	konevcny2016federated}. 
	 	Other approaches for low-communication training focus on sparsification of gradients, either by thresholding small entries or by random sampling~\cite{
	 	strom2015scalable,
	 	mania2015perturbed,
	 	suresh2016distributed,
	 	leblond2016asaga,
	 	aji2017sparse,
	 	konevcny2016randomized,
	 	lin2017deep,
	 	chen2017adacomp,
	 	renggli2018sparcml,
	 	tsuzuku2018variance,
	 	wang2018atomo,
	 	vogels2019powersgd}.  
	 
	 However, the proposed communication-efficient training techniques via gradient compression usually suffer from some of the following drawbacks: (i) The computation cost for gradient compression (\eg sparsification or quantization) can be high. 
	 For instance, \textsc{Atomo} \citep{wang2018atomo} requires to compute gradient factorizations using SVD for every single batch, which can be computationally expensive for large-scale models. 
	 (ii) Existing gradient compression methods either do not fully utilize the full gradients ~\citep{alistarh2017qsgd,wen2017terngrad,bernstein2018signsgd,wang2018atomo} or require additional memory. For example, the ``\textit{error feedback}" scheme~\citep{seide20141,stich2018sparsified,karimireddy2019error} utilizes stale gradients aggregated in memory for future iterations, but requires storing additional information proportional to the model size. (iii) Significant implementation efforts are required to incorporate an existing gradient compression technique within high-efficiency distributed training APIs in current deep learning frameworks \eg \texttt{DistributedDataParallel} (DDP) in PyTorch. 
	 
	 Due to the above shortcomings of current communication-efficient techniques, it is of interest to explore the feasibility of incorporating elements of the gradient compression step into the model architecture itself. If this is feasible, then communication efficiency can be attained at no extra cost. 
	 In this work, we take a first step towards bypassing the gradient compression step via training low-rank, pre-factorized deep network, starting from full-rank counterparts. We observe that training low-rank models from scratch incurs non-trivial accuracy loss. To mitigate that loss, instead of starting from a low-rank network, we initialize at a full-rank counterpart. We train for a small fraction, \eg 10\% of total number epochs, with the full-rank network, and then convert to a low-rank counterpart. To obtain such a low-rank model we apply SVD on each of the layers. After the SVD step, we use the remaining 90\% of the training epochs to fine-tune this low-rank model.
	 The proposed method bares similarities to the ``\textit{Lottery Ticket Hypothesis}" (LTH) \cite{frankle2018lottery}, in that we find  ``winning tickets" within full-rank/dense models, but without the additional burden of ``winning the lottery''. Winning tickets seem to be in abundance once we seek models that are sparse in their spectral domain.
	 
	 \vspace{-2 mm}
	 \paragraph{Our contributions.} In this work, we propose \methodname{}, a computation and communication efficient distributed training framework. \methodname{} takes any deep neural network architecture and finds a pre-factorized low-rank representation. \methodname{} then trains the pre-factorized low-rank network to achieve both computation and communication efficiency, instead of explicitly compressing gradients. \methodname{} supports several types of architectures including fully connected (FC), convolutional neural nets (CNNs), LSTMs, and Transformer networks~\citep{vaswani2017attention}. As \methodname{} manipulates the model architectures instead of their gradients, it is directly compatible with all SOTA distributed training frameworks, \eg PyTorch DDP and BytePS~\citep{jiang2020unified}.
	 

    \begin{figure}[htp]
    	\centering
    	\includegraphics[width=0.35\textwidth]{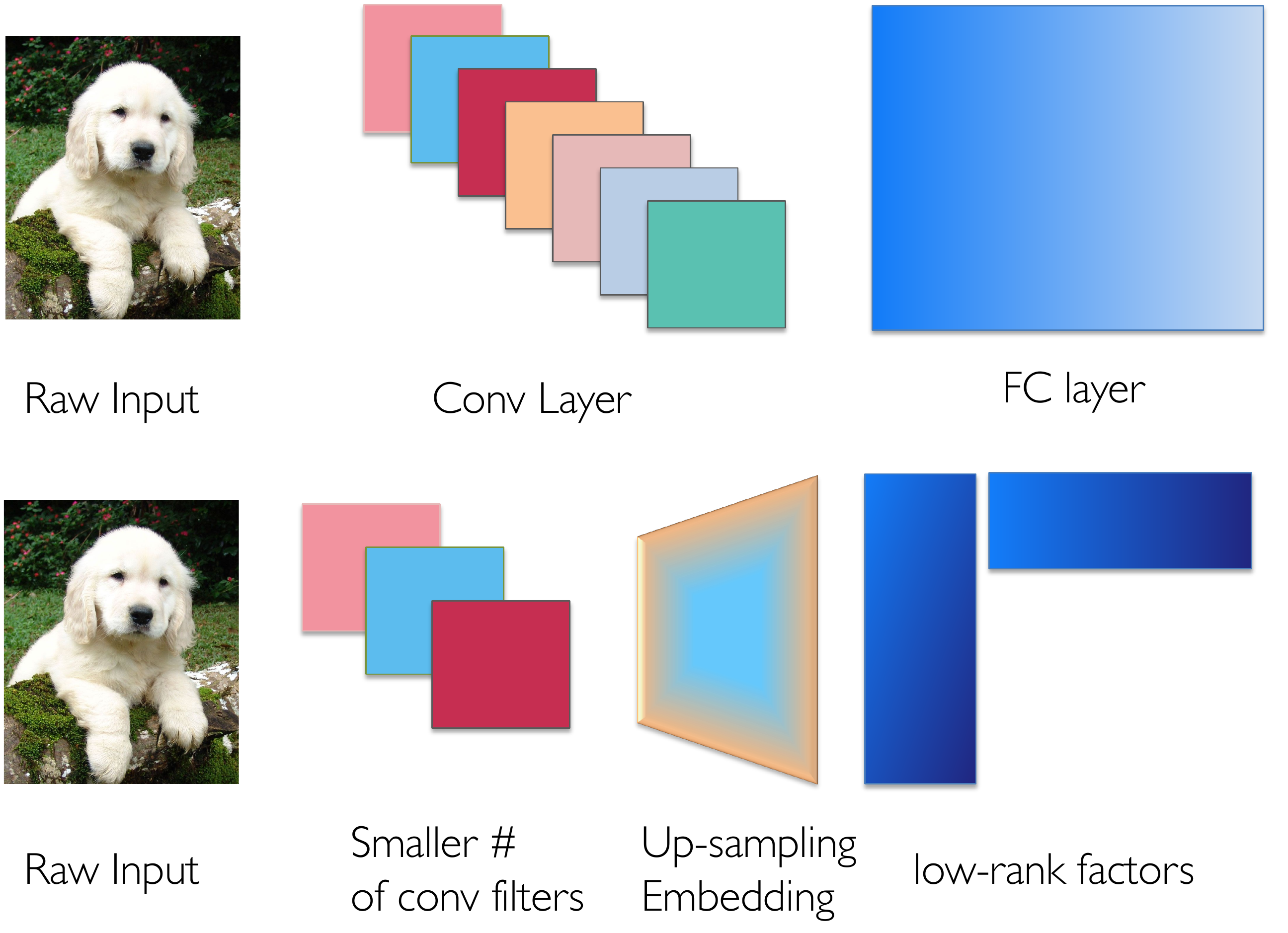}
    	\vspace{-2 mm}
    	\caption{ We propose to replace fully connected layers  represented by a matrix $W$, by a set of trainable factors $UV^T$, and represent each of the $N$ convolutional filters of each conv layer as a linear combination of $\frac{N}{R}$ filters. This latter operation can be achieved by using fewer filters per layer, and then applying a trainable up-sampling embedding to the output channels.
    	}
    	\label{fig:lowrankDNN}
    \vspace{-4 mm}
    \end{figure}
    
    We further observe that direct training of those pre-factorized low-rank deep networks leads to non-trivial accuracy loss, especially for large-scale machine learning tasks, \eg ImageNet \cite{deng2009imagenet}. We develop two techniques for mitigating this accuracy loss: (i) a \textit{ hybrid architecture} and (ii) \textit{vanilla warm-up training}. The effectiveness of these two techniques is justified via extensive experiments. 
    
    We provide  experimental results over real distributed systems and large-scale vision and language processing tasks. We compare \methodname{} against a wide range of SOTA baselines: (i) communication-efficient distributed training methods \eg \textsc{PowerSGD}~\cite{vogels2019powersgd} and \textsc{Signum}~\cite{bernstein2018signsgd}; (ii) structured pruning methods, \eg the \textit{Early Bird Ticket} (EB Train)~\cite{you2019drawing}; and model sparsification method, \eg the iterative pruning algorithm in LTH~\citep{frankle2018lottery}. Our experimental results indicate that \methodname{} achieves better model training efficiency compared to \textsc{PowerSGD}, \textsc{signum}, and  LTH models. \methodname{} also leads to smaller and more accurate model compared to EB Train. We further show that the performance of \methodname{} remains stable under \textit{mixed-precision training}.
    
    \vspace{-2 mm}
	\paragraph{Related work.} \methodname{} is closely related to the work on communication-efficient distributed training methods. To reduce the communication cost in distributed training, the related literature has developed several methods for gradient compression. Some of the methods use quantization over the gradient elements ~\citep{seide20141,alistarh2017qsgd,wen2017terngrad,lin2017deep,luo2017thinet,bernstein2018signsgd,tang2019doublesqueeze,wu2018error}. Other methods study sparsifying the gradients in the element-wise or spectral domains~\citep{lin2017deep,wang2018atomo,stich2018sparsified,vogels2019powersgd}. It has also been widely observed that adopting the ``\textit{error feedback}" scheme is generally helpful for gradient compression methods to achieve better final model accuracy~\citep{stich2018sparsified,wu2018error,karimireddy2019error,vogels2019powersgd}. Compared to the previously proposed gradient compression methods, \methodname{}  merges the gradient compression into model training, thus achieves communication-efficiency at no extra cost.
	
    \methodname{} is also closely related to model compression. Partially initialized by \textit{deep compression}~\citep{han2015deep}, a lot of research proposes to remove the redundant weights in the trained neural networks. The trained neural networks can be compressed via model weight pruning~\citep{li2016pruning,wen2016learning,hu2016network,zhu2017prune,he2017channel,yang2017designing,liu2018rethinking,yu2018nisp,yu2018slimmable}, quantization ~\citep{rastegari2016xnor,zhu2016trained,hubara2016binarized,wu2016quantized,hubara2017quantized,zhou2017incremental}, and low-rank factorization~\citep{xue2013restructuring,sainath2013low,jaderberg2014speeding,wiesler2014mean,konevcny2016federated}. Different from the model compression methods, \methodname{} proposes to train the factorized networks, which achieves better overall training time, rather than compressing the model after fully training it.
    
    Finally, our work is also related to efficient network architecture design, where the network layers are re-designed to be smaller, more compact, and more efficient ~\citep{iandola2016squeezenet,chen2016eyeriss,zhang2018shufflenet,tan2019efficientnet,howard2017mobilenets,chollet2017xception,lan2019albert,touvron2020fixing,waleffe2020principal}. The most related low-rank efficient training framework to \methodname{} is the one proposed in~\citep{ioannou2015training}, where a pre-factorized network is trained from scratch. However, we demonstrate that training the factorized network from scratch leads to non-trivial accuracy loss. In \methodname{}, we propose to warm-up the low-rank model via factorizing a partially trained full-rank model. Our extensive experiments indicate that \methodname{} achieves significantly higher accuracy compared to training the factorized network from scratch. Moreover, \citep{ioannou2015training} only studies low-rank factorizations for convolutional layers, whereas \methodname{} supports FC, CNN, LSTM, and Transformer layers.
		
    \vspace{-2 mm}
\section{\textsc{PufferFish}: effective deep factorized network training}\label{sec:pufferfish}

In the following subsections, we discuss how model factorization is implemented for different model architectures.

\subsection{Low-rank factorization for FC layers}
For simplicity, we discuss a 2-layer FC network that can be represented as $h(x) = \sigma(W_1\sigma(W_2x))$ where $W_l, \forall l \in \{1, 2\}$ are  weight matrices, $\sigma(\cdot)$ is an arbitrary activation function, and $x$ is the input data point.
We propose to pre-factorize the matrices $W_l$ into $U_l V_l^T$ where the factors are of significantly smaller dimensions while also reducing the computational complexity of the full-rank FC layer.
\vspace{-2 mm}
\subsection{Low-rank factorization for convolution layers} 
\paragraph{Basics on convolution layers.} The above low-rank factorization strategy extends to convolutional layers (see Fig.~\ref{fig:lowrankDNN} for a sketch). 
In a convolution layer, a $c_{\text{in}}$-channel input image of size $H\times W$ pixels is convolved with $c_{\text{out}}$ filters of size $c_{\text{in}}\times k\times k$ to create a $c_{\text{out}}$-channel output feature map. Therefore, the computational complexity for the convolution of the filter with a $c_{\text{in}}$-channel input image is
$\mathcal{O}(c_{\text{in}}c_{\text{out}}k^2HW)$. In what follows, we describe schemes for modifying the architecture of the convolution layers via low-rank factorization to reduce computational complexity and the number of parameters. The idea is to replace vanilla (full-rank) convolution layers with
factorized versions. These factorized filters amount to the same number of convolution filters, but are constructed through linear combinations of a sparse, \ie low-rank filter basis.
\vspace{-2mm}
\paragraph{Factorizing a convolution layer.} 
For a convolution layer with dimension $W\in \mathbb{R}^{c_\text{in} \times c_\text{out} \times k \times k}$ where $c_\text{in}$ and $c_\text{out}$ are the number of input and output channels and $k$ is the size of the convolution filters, \eg $k=3$ or $5$. Instead of factorizing the 4D weight of a convolution layer directly, we consider factorizing the unrolled 2D matrix. Unrolling the 4D tensor $W$ leads to a 2D matrix with shape $W_{\text{unrolled}} \in \mathbb{R}^{c_\text{in}k^2 \times c_\text{out}}$ where each column represents the weight of a vectorized convolution filter. The rank of the unrolled matrix is determined by $\min\{c_{\text{in}}k^2,c_{\text{out}}\}$. Factorizing the unrolled matrix returns $U \in \mathbb{R}^{c_\text{in}k^2\times r}$, $V^\top \in \mathbb{R}^{r \times c_\text{out}}$, \ie $W_{\text{unrolled}} \approx UV^\top$. Reshaping the factorized $U, V^\top$ matrices back to 4D filters leads to $U \in \mathbb{R}^{c_{\text{in}} \times r \times k \times k}, V^\top \in \mathbb{R}^{r \times c_{\text{out}}}$. Therefore, factorizing a convolution layer returns a thinner convolution layer $U$ with width $r$, \ie the number of convolution filters, and a linear projection layer $V^\top$. In other words, the full-rank original convolution filter bank is approximated by a linear combination of $r$ basis filters. The $V^\top$s can also be represented by a $1\times 1$ convolution layer, \eg $V^\top_l \in \mathbb{R}^{r \times c_{\text{out}} \times 1 \times 1}$, which is more natural for computer vision tasks as it operates directly on the spatial domain~\cite{lin2013network}. In \methodname{}, we use the $1\times1$ convolution for all $V^\top_l$ layers in the considered CNNs. One can also use tensor decomposition, \eg the Tucker decomposition to directly factorize the 4D tensor weights~\cite{tucker1966some}. In this work, for simplicity, we do not consider tensor decompositions. 
\vspace{-2 mm}
\subsection{Low-rank factorization for LSTM layers} 
LSTMs have been proposed as a means to mitigate the ``vanishing gradient'' issue of traditional RNNs~\cite{hochreiter1997long}. The forward pass of an LSTM is as follows
\begin{align} 
i_t &= \sigma(W_{ii} x_t + b_{ii} + W_{hi} h_{t-1} + b_{hi})\nonumber \\
f_t &= \sigma(W_{if} x_t + b_{if} + W_{hf} h_{t-1} + b_{hf})\nonumber \\
g_t &= \tanh(W_{ig} x_t + b_{ig} + W_{hg} h_{t-1} + b_{hg}) \label{eq:lstm-rule}\\
o_t &= \sigma(W_{io} x_t + b_{io} + W_{ho} h_{t-1} + b_{ho}) \nonumber\\
c_t &= f_t \odot c_{t-1} + i_t \odot g_t \nonumber \\
h_t &= o_t \odot \tanh(c_t)\nonumber.
\end{align}
$h_t, c_t, x_t$ represent the hidden state, cell state, and  input at time $t$ respectively. $h_{t-1}$ is the hidden state of the layer at time $t-1$. $i_t, f_t, g_t, o_t$ are the input, forget, cell, and output gates, respectively. $\sigma(\cdot)$ and $\odot$ denote the sigmoid activation function and the Hadamard product, respectively. The trainable weights are the matrices $W_{i\cdot} \in \mathbb{R}^{h\times d}, W_{h\cdot}\in \mathbb{R}^{h\times h}$, where $d$ and $h$ are the embedding and hidden dimensions. Thus, similarly to the low-rank FC layer factorization, the factorized LSTM layer is represented by
\begin{align} 
i_t &= \sigma(U_{ii}V^\top_{ii} x_t + b_{ii} + U_{hi}V^\top_{hi} h_{t-1} + b_{hi})\nonumber \\
f_t &= \sigma(U_{if}V^\top_{if} x_t + b_{if} + U_{hf}V^\top_{hf} h_{t-1} + b_{hf})\nonumber \\
g_t &= \tanh(U_{ig}V^\top_{ig} x_t + b_{ig} + U_{hg}V^\top_{hg} h_{t-1} + b_{hg})\label{eq:lr-lstm-rule} \\
o_t &= \sigma(U_{io}V^\top_{io} x_t + b_{io} + U_{ho}V^\top_{ho} h_{t-1} + b_{ho})\nonumber \\
c_t &= f_t \odot c_{t-1} + i_t \odot g_t\nonumber \\
h_t &= o_t \odot \tanh(c_t)\nonumber.
\end{align}
\vspace{-6 mm}
\subsection{Low-rank network factorization for Transformer}\label{sec:low-rank-transformer}
A Transformer layer consists of a stack of encoders and decoders~\cite{vaswani2017attention}. Both encoder and decoder contain three main building blocks, \ie the \textit{multi-head attention} layer, \textit{position-wise feed-forward networks} (FFN), and \textit{positional encoding}. A $p$-head attention layer learns $p$ independent attention mechanisms on the input key ($K$), value ($V$), and queries ($Q$) of each input token: 
\begin{align*}
\text{MultiHead}(Q, K, V)&=\text{Concat}(\text{head}_1,\cdots, \text{head}_p)W^O\\
\text{where head}_i&=\text{Attention}(QW^Q_i, KW^K_i, VW^V_i). \label{eq:scaled-dot-product-attention}
\end{align*}
In the above, $W_i^Q, W_i^K, W_i^V, i \in \{1, \cdots, p\}$ are trainable weight matrices. The particular attention, referred to as ``scaled dot-product attention", is used in Transformers, \ie $\text{Attention}(\tilde Q, \tilde K, \tilde V) = \text{softmax}\bigg(\frac{\tilde Q \tilde K^\top}{\sqrt{d}}\bigg)\tilde V$ where $\tilde Q = Q W^Q_i, \tilde K = K W^K_i, \tilde V = V W^V_i$.
$W^O$ projects the output of the multi-head attention layer to match the embedding dimension. Following~\cite{vaswani2017attention}, we assume the projected key, value, and query are embedded to $pd$ dimensions, and are projected to $d$ dimensions in the attention layer. 
In Transformer, a sequence of $N$ input tokens are usually batched before passing to the model where each input token is embedded to a $pd$ dimensional vector. Thus, dimensions of the inputs are $Q, K, V \in \mathbb{R}^{N\times pd}$.  The learnable weight matrices are $W_i^Q, W_i^K, W_i^V\in \mathbb{R}^{pd\times d}, W^O \in \mathbb{R}^{pd\times pd}$. The FFN in Transformer consists of two learnable FC layers: $\text{FFN}(x) = \text{max}(0, x W_1 + b_1)W_2 + b_2$ where $W_1 \in \mathbb{R}^{pd\times 4pd}, W_2 \in \mathbb{R}^{4pd \times pd}$ (the relationships between the notations in our paper and the original Transformer paper~\cite{vaswani2017attention} are $pd = d_{\text{model}}, d = d_k = d_v$, and $d_{ff} = 4pd$). 

In \methodname{}, we factorize all learnable weight matrices in the multi-head attention and the FFN layers. We leave the positional encoding as is, since there are no trainable weights. For the bias term of each layer and the ``\textit{Layer Normalization}" weights, we use the vanilla weights directly, as they are represented by vectors.
\begin{table}[ht]
    \vspace{-4mm}
	\caption{The number of parameters and computational complexities for full-rank and low-rank FC, convolution, LSTM, and the Transformer layers where $m$, $n$ are the dimensions of the FC layer and $c_{\text{in}}, c_{\text{out}}, k$ are the input, output dimensions, and kernel size respectively. $h, d$ denote the hidden and embedding dimensions in the LSTM layer. $N,p,d$ denote the sequence length, number of heads, and embedding dimensions in the Transformer. $r$ denotes the rank of the factorized low-rank layer we assume to use. }
	\label{table:complexities}
	\begin{center}
	  \scriptsize{
		\begin{tabular}{ccc}
		\toprule \textbf{Networks}
		& \# Params. &  Computational Complexity \bigstrut\\
		\midrule
		Vanilla FC & $m \times n$ & $\mathcal{O}(m n)$ \bigstrut\\
		Factorized FC & $r(m + n)$ & $\mathcal{O}(r(m + n))$ \bigstrut\\
		Vanilla Conv. & $c_{\text{in}}\times c_{\text{out}} \times k^2$ & $\mathcal{O}(c_{\text{in}} c_{\text{out}} k^2 HW)$ \bigstrut\\
		Factorized Conv. & $c_{\text{in}}rk^2+rc_{\text{out}}$ & $\mathcal{O}(rc_{\text{in}}k^2 HW+rHW c_{\text{out}})$ \bigstrut\\
		Vanilla LSTM & $4(dh+h^2)$ & $\mathcal{O}(dh+h^2)$ \bigstrut\\
		Factorized LSTM & $4dr+12hr$ & $\mathcal{O}(dr+hr)$ \bigstrut\\
		Vanilla Attention & $4p^2d^2$ & $\mathcal{O}(Np^2d^2+N^2d)$ \bigstrut\\
		Factorized Attention & $(3p+5)prd$ & $\mathcal{O}\big(rpdN +N^2 d \big)$ \bigstrut\\
		Vanilla FFN & $8 p^2 d^2$ & $\mathcal{O}\big(p^2 d^2 N\big)$ \bigstrut\\
		Factorized FFN & $10pdr$ & $\mathcal{O}\big(r p dN\big)$ \bigstrut\\
		\bottomrule
		\end{tabular}}%
	\vspace{-6mm}
	\end{center}
\end{table}
\vspace{-4 mm}
\subsection{Computational complexity and model size} A low-rank factorized network enjoys a smaller number of parameters and lower computational complexity. Thus, both the computation and communication efficiencies are improved, as the amount of communication is proportional to the number of parameters. We summarize the computational complexity and the number of parameters in the vanilla and low-rank FC, convolution, LSTM, and the Transformer layers in Table~\ref{table:complexities}. We assume the FC layer has shape $W_{FC} \in \mathbb{R}^{m \times n}$, the convolution layer has shape $W_{\text{Conv}} \in \mathbb{R}^{c_{\text{in}} \times c_{\text{out}} \times k \times k}$, the LSTM layer has shape $W_i\in \mathbb{R}^{4h\times d}; W_h \in \mathbb{R}^{4h\times h}$ (where $W_i$ and $W_h$ is the concatenated input-hidden and hidden-hidden weight matrices), and the shapes of the model weights in the encoder of a Transformer follow the discussion in Section~\ref{sec:low-rank-transformer}. For Transformers, we show the computational complexity of a single encoder block. We assume the low-rank layers have rank $r$. As the computation across the $p$ heads can be done in parallel, we report the computational complexity of a single attention head. Note that for the LSTM layer, our complexity analysis assumes the low-rank layer uses the same rank for the input-hidden weights $W_{i\cdot}$ and the hidden-hidden weights $W_{h\cdot}$. Similarly, for the Transformer layer, we assume the low-rank layer uses the same rank $r$ for all $W_i^Q, W_i^K, W_i^V, W^O$. Further details can be found in the Appendix.
\vspace{0.1cm}
\section{Strategies for mitigating  accuracy loss}
\vspace{-0.1cm}
    \begin{figure}[ht]
        \vspace{-4 mm}
    	\centering
    	\subfigure[VGG-11 on CIFAR-10]{\includegraphics[width=0.22\textwidth]{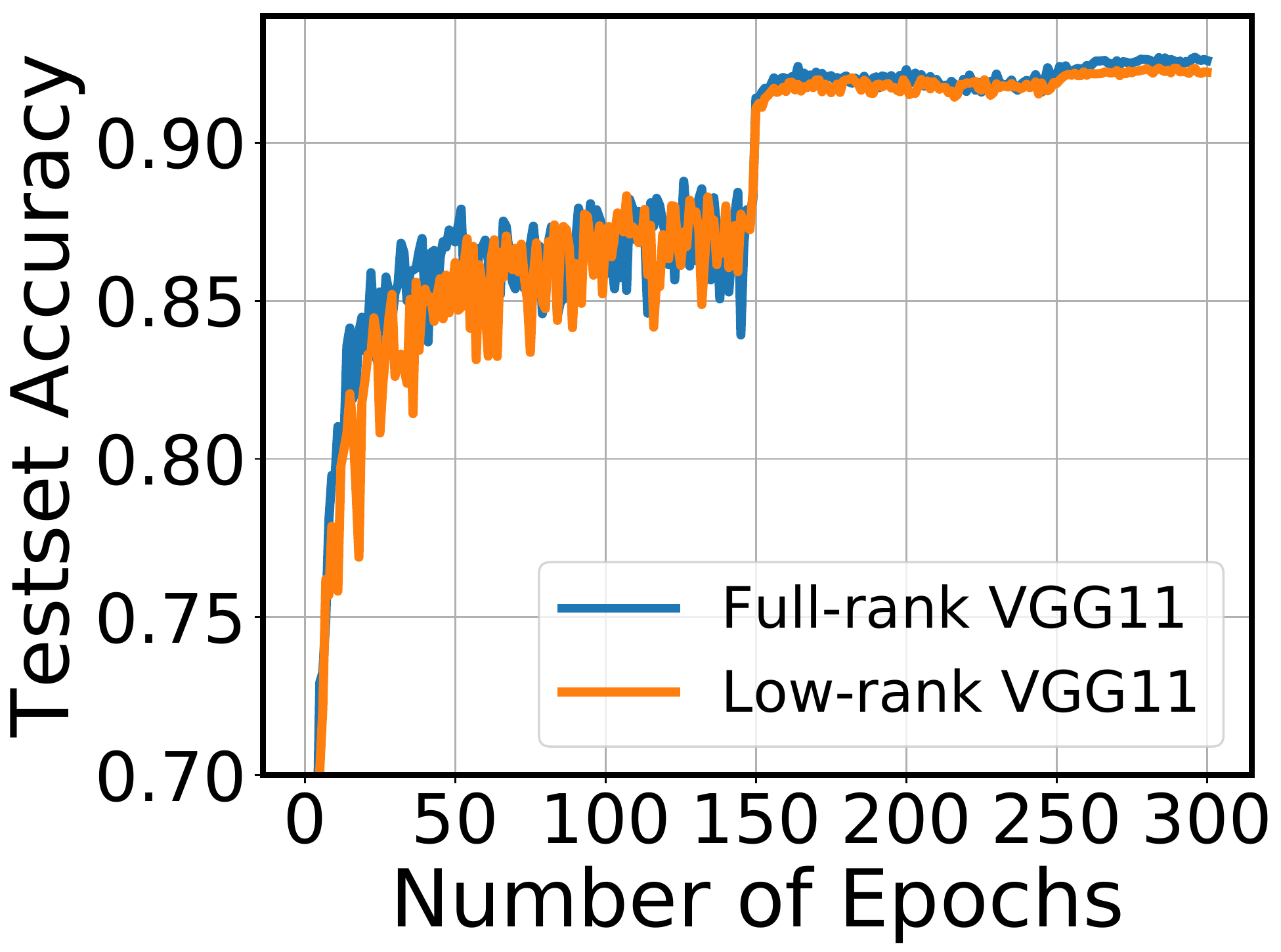}}
    	\subfigure[ResNet-50 on ImageNet]{\includegraphics[width=0.22\textwidth]{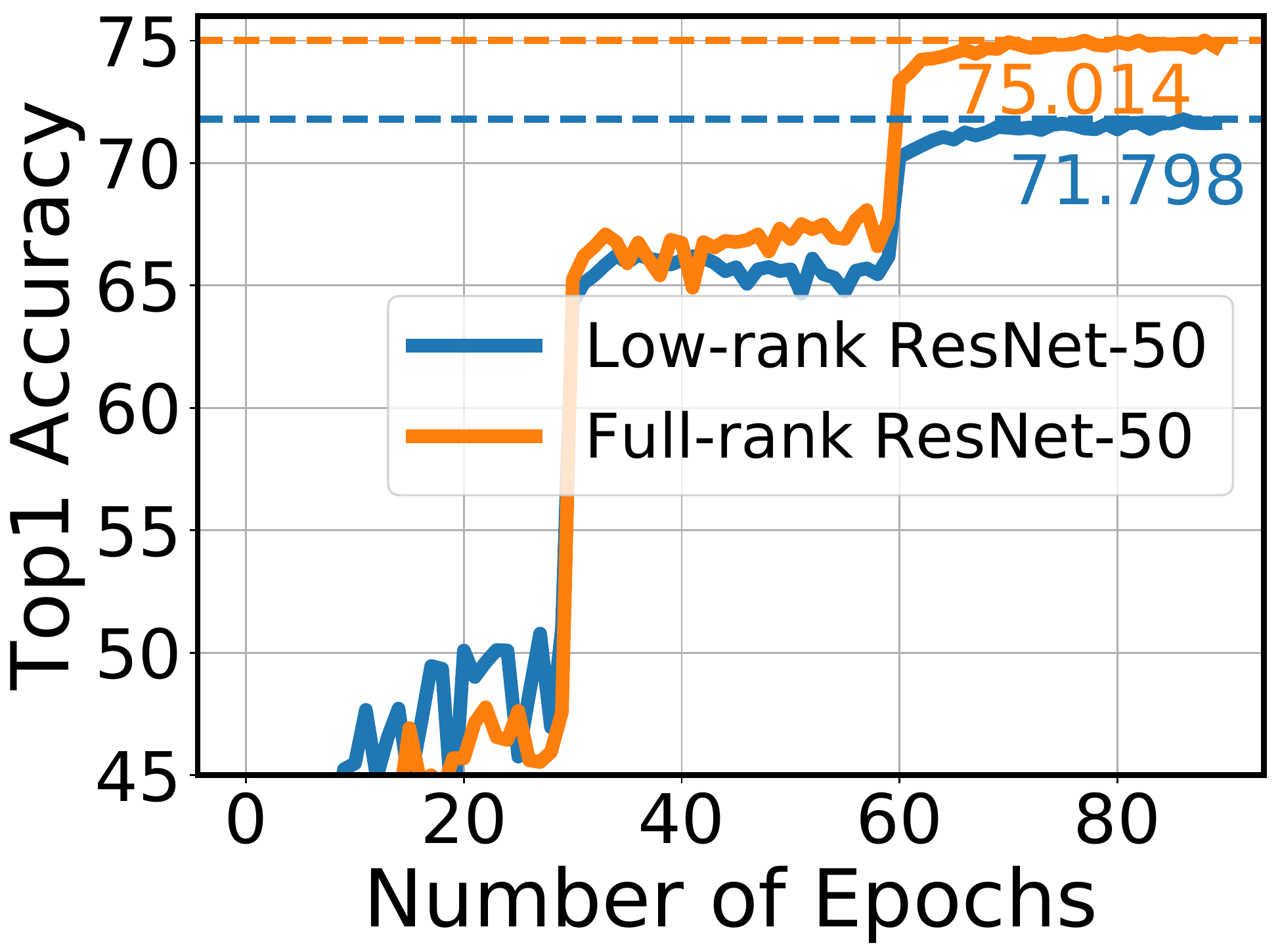}}
    	\vspace{-4 mm}
    	\caption{Model convergence comparisons between vanilla models and \methodname{} factorized models: (a) low-rank VGG-11 over the CIFAR-10 dataset; (b)  ResNet-50 over the ImageNet dataset. For the low-rank networks, all layers except for the first convolution and the very last FC layer are factorized with a fixed rank ratio at $0.25$.
    	}
    	\label{fig:lr-vgg}
    	\vspace{-4 mm}
    \end{figure}
    \begin{figure}[ht]
    	\centering
    	\subfigure[Hybrid network]{\includegraphics[width=0.215\textwidth]{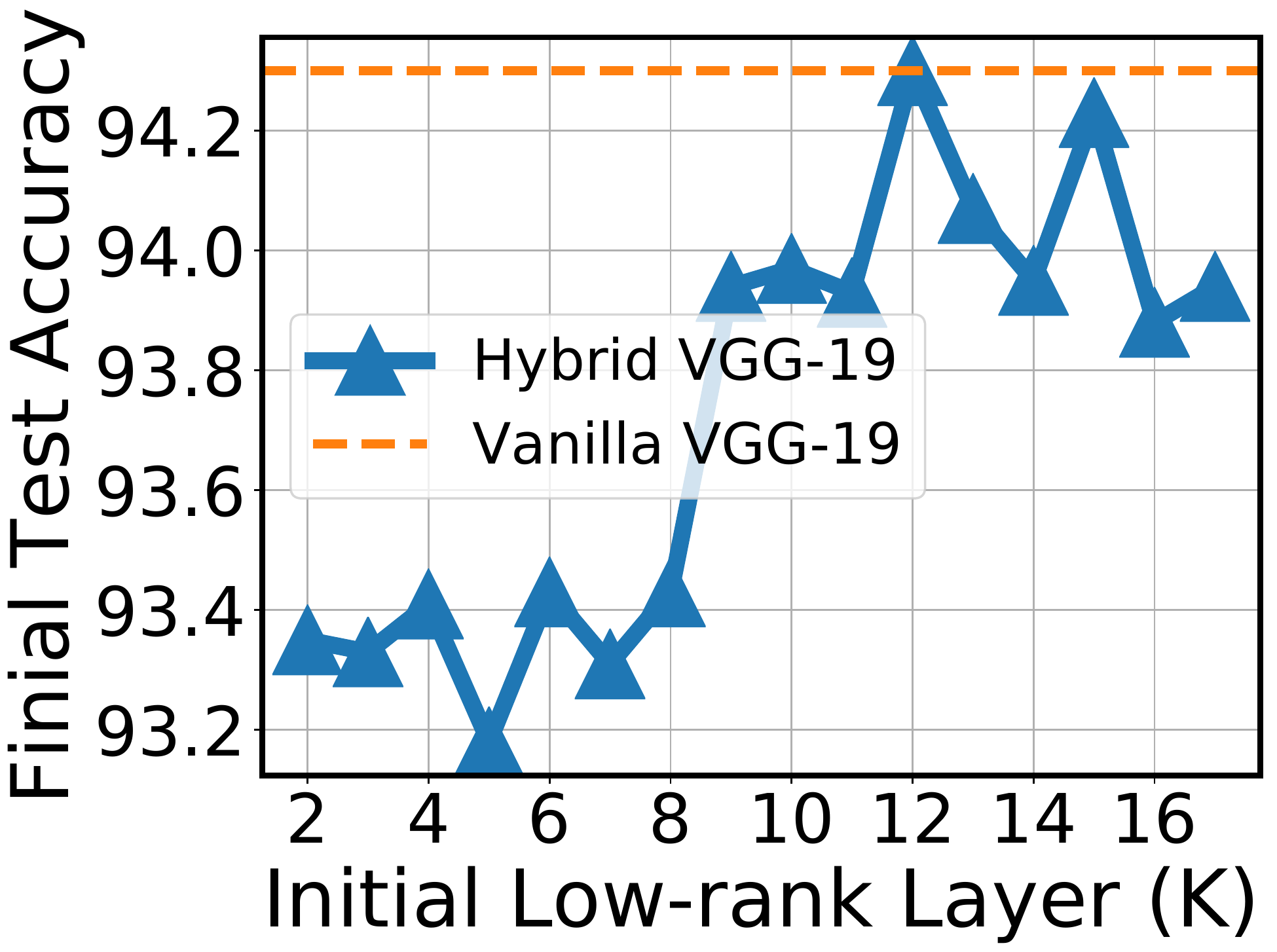}\label{fig:effect-hybrid}}
    	\subfigure[Vanilla warm-up training]{\includegraphics[width=0.255\textwidth]{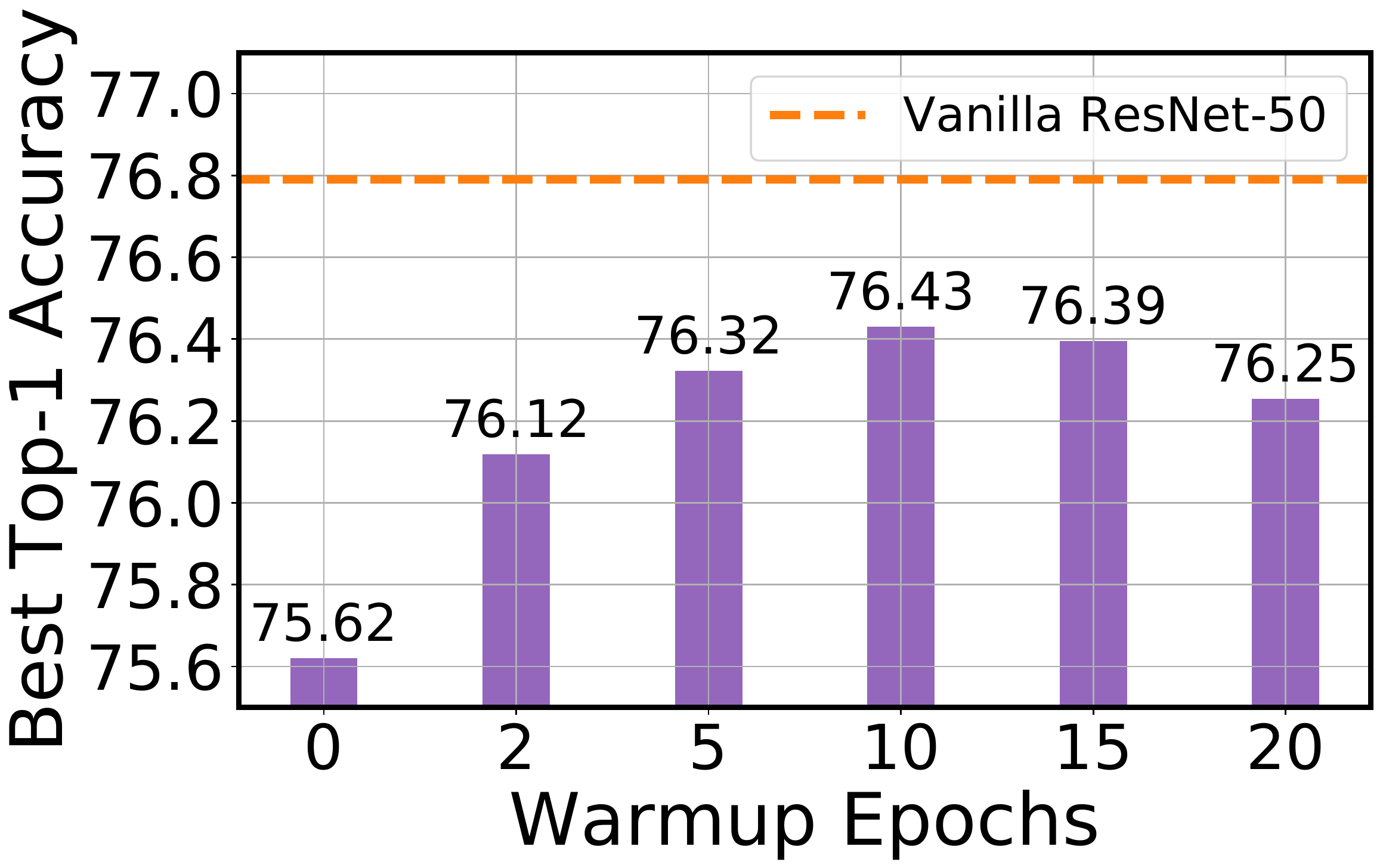}\label{fig:effect-fr-warmup}}
    	\vspace{-4 mm}
    	\caption{The effect of the test accuracy loss mitigation methods in \methodname{}: (a) \textbf{Hybrid network}: The final test accuracy of the hybrid VGG-19 architectures with various initial low-rank layer indices ($K$) over the CIFAR-10 dataset. (b) \textbf{Vanilla warm-up training}: The final top-1 accuracy of the hybrid-ResNet-50 architecture trained on the ImageNet dataset under the different number of vanilla warm-up epochs: $\{2, 5, 10, 15, 20\}$.
    	}
    	\label{fig:lr-vgg}
    	\vspace{-6 mm}
    \end{figure}
    
    In this section, we showcase that training low-rank models from scratch leads to an accuracy loss. Interestingly, this loss can be mitigated by balancing the degree of factorization across layers, and by using a short full-rank warm-up training phase used to initialize the factorized model.
    
We conduct an experimental study on a version of \methodname{} where every layer of the network is factorized except for the first convolution layer and the last FC layer. On a relatively small task, \eg VGG-11 on CIFAR-10, we observe that \methodname{} only leads to $\sim 0.4\%$ accuracy loss (as shown in Figure~\ref{fig:lr-vgg}) compared to the vanilla VGG-19-BN. However, for ResNet-50 on the ImageNet dataset, a $\sim 3\%$ top-1 accuracy loss of \methodname{} is observed. To mitigate the accuracy loss of the factorized networks over the large-scale ML tasks, we propose two methods, \ie (i) \textit{hybrid network architecture} and (ii) \textit{vanilla warm-up training}. We then discuss each method separately.
\vspace{-2 mm}
\paragraph{Hybrid network architecture.} In \methodname{}, the low-rank factorization aims at approximating the original network weights, \ie $W_l \approx U_l V^\top_l$ for layer $l$, which inevitably introduces approximation error. Since the approximation error in the early layers can be accumulated and propagated to the  later layers, a natural strategy to mitigate the model accuracy loss is to only factorize the later layers. Moreover, for most of CNNs, the number of parameters in later layers dominates the entire network size. Thus, factorizing the later layers does not sacrifice the degree of model compression we can achieve. 
Specifically, for an $L$ layer network $\{W_1, W_2, \cdots, W_L \}$, factorizing every layer leads to $\{U_1, V_1^\top, U_2, V^\top_2, \cdots, U_L, V^\top_L \}$. In the hybrid network architecture, the first $K-1$ layers are not factorized, \ie $\{W_1, W_2, \cdots, W_{K-1}, U_{K}, V^\top_{K}, \cdots, U_L, V^\top_L \}$ where we define $K$ as the index of the first low-rank layer in a hybrid architecture. We treat $K$ as a hyper-parameter, which balances the model compression ratio and the final model accuracy. In our experiments, we tune $K$ for all models. The effectiveness of the hybrid network architecture is shown in Figure~\ref{fig:effect-hybrid}, from which we observe that 
the hybrid VGG-19 with $K=9$ mitigates $\sim0.6\%$ test accuracy loss.
\vspace{-2 mm}
\begin{algorithm}[t]
\small
	\SetKwInOut{Input}{Input}
	\SetKwInOut{Output}{Output}
	\Input{Randomly initialized weights of vanilla $N$-layer architectures $\{W_{1}, W_{2}, \ldots ,W_{L}\}$, and the associated weights of hybrid $N$-layer architecture $\{W_{1}, W_{2}, \ldots, W_{K-1}, U_{K}, V^\top_{K}, \ldots ,U_{L}, V^\top_{L}\}$, the entire training epochs $E$, the vanilla warm-up training epochs $E_{wu}$, and learning rate schedule $\{\eta_t\}^E_{t=1}$}
	\Output{Trained hybrid $L$-layer architecture weights $\{\hat W_{1}, \hat W_{2}, \ldots, \hat W_{K-1}, \hat U_{K}, \hat V^\top_{K}, \ldots , \hat U_{L}, \hat V^\top_{L}\}$}

    \For{$t \in \{1,\ldots, E_{wu}\}$}{
    Train $\{W_{1}, W_{2}, \ldots ,W_{L}\}$ with learning rate schedule $\{\eta_t\}_{t=1}^{E_{wu}}$ \tcp*{vanilla warm-up training}
    }
    \For{$l \in \{1,\ldots, L\}$}{
        \uIf{$l < K$}{
            copy the partially trained $W_l$ weight to the hybrid network;    
        }
        \Else{
            $\tilde U_l \Sigma_l \tilde V_l^\top = \text{SVD}(W_l)$  \tcp*{Decomposing the vanilla warm-up trained weights}
            $U_l = \tilde U_l \Sigma_l^{\frac{1}{2}}, V_l^\top = \Sigma^{\frac{1}{2}}\tilde V_l^\top$
        }
    }
    \For{$t \in \{E_{wu}+1,\ldots, E\}$}{
    Train the hybrid network weights, \ie $\{W_{1}, W_{2}, \ldots, W_{K-1}, U_{K}, V^\top_{K}, \ldots , U_{L}, V^\top_{L}\}$ with learning rate schedule $\{\eta_t\}_{t=E_{wu}}^{E}$ \tcp*{consecutive low rank training}
    }    
\caption{\methodname{} Training Procedure}
\label{alg:pufferfish}
\end{algorithm}
\paragraph{Vanilla warm-up training.} It has been widely observed that epochs early in training are critical for the final model accuracy~\cite{jastrzebski2020break,keskar2016large,achille2018critical,leclerc2020two,agarwal2020accordion}. For instance, sparsifying gradients in early training phases can hurt the final model accuracy~\citep{lin2017deep}. Similarly, factorizing the vanilla model weights in the very beginning of the training procedure can also lead to accuracy loss, which may be impossible to mitigate in later training epochs. It has also been shown that good initialization strategies play a significant role in the final model accuracy~\citep{zhou2020go}. 

In this work, to mitigate the accuracy loss, we propose to use the partially trained vanilla, full-rank model weights to initialize the low-rank factorized network. We refer to this as ``\textit{vanilla warm-up training}". We train the vanilla model for a few epochs ($E_{wu}$) first. Then, we conduct truncated matrix factorization (via truncated SVD) over the partially trained model weights to initialize the low-rank factors. For instance, given a partially trained FC layer $W^{(l)}$, we deploy SVD on it such that we get $\tilde U\Sigma \tilde V^\top$. After that the $U$ and $V^\top$ weights we introduced in the previous sections can be found by $U = \tilde U \Sigma^{\frac{1}{2}}, V^\top = \Sigma^{\frac{1}{2}}\tilde V^\top$. For convolution layer $W \in \mathbb{R}^{c_{\text{in}}\times c_{\text{out}} \times k\times k}$, we conduct SVD over the unrolled 2D matrix $W_{\text{unrolled}} \in \mathbb{R}^{c_\text{in}k^2 \times c_\text{out}}$, which leads to $U\in \mathbb{R}^{c_{\text{in}}k^2 \times r}, V^\top \in \mathbb{R}^{r \times c_{\text{out}}}$ where reshaping $U, V$ back to 4D leads to the desired initial weights for the low-rank layer, \ie $U\in \mathbb{R}^{r\times c_{\text{out}} \times k \times k}, V^\top \in \mathbb{R}^{r \times c_{\text{out}} \times 1 \times 1}$. For the Batch Normalization layers (BNs) \cite{ioffe2015batch} we simply extract the weight vectors and the collected running statistics, \eg the \textit{running mean and variance}, for initializing the low-rank training. We also directly take the bias vector of the last FC layer. \methodname{} then finishes the remaining training epochs over the factorized hybrid network initialized with vanilla warm-up training. 

Figure~\ref{fig:effect-fr-warmup} provides an experimental justification on the effectiveness of vanilla warm-up training where we study a hybrid ResNet-50 trained on the ImageNet dataset. The results indicate that vanilla warm-up training helps to improve the accuracy of the factorized model. Moreover, a carefully tuned warm-up period of $\hat E_{wu}$ also plays an important role in the final model accuracy. Though SVD is computationally heavy, \methodname{} only requires to conduct the SVD \textbf{once} throughout the entire training. We benchmark the SVD cost for all experimented models, which indicate the SVD runtime is comparatively small, \eg on average, it only costs $2.29$ seconds for ResNet-50. A complete study on the SVD factorization overheads can be found in the Appendix.

\vspace{-0.5cm}
\paragraph{Last FC layer.} 
The very last FC layer in a neural network can be viewed as a linear classifier over the features extracted by the previous layers. 
In general, its rank is equal to the number of classes in predictve task at hand. Factorizing it below the number of classes, will increase linear dependencies, and may further increase the approximation error. Thus, \methodname{} does not factorize it. 

Putting all the techniques we discussed in this section together, the training procedure of \methodname{} is summarized in Algorithm~\ref{alg:pufferfish}. 
\vspace{-2 mm}
\section{Experiments}\label{sec:experiment}
We conduct extensive experiments to study the effectiveness and scalability of \methodname{} over various computer vision and natural language processing tasks, across real distributed environments. We also compare \methodname{} against a wide range of baselines including: (i) \textsc{PowerSGD}, a low-rank based, gradient compression method that achieves high compression ratios~\citep{vogels2019powersgd}; (ii) \textsc{Signum} a gradient compression method that only communicates the sign of the local momentum~\citep{bernstein2018signsgd,bernstein2018signsgd2}; 
(iii) The ``early bird'' structured pruning method \textit{EB Train}~\citep{you2019drawing}; and (iv) The LTH sparsification method (referred to as LTH for simplicity)~\citep{frankle2018lottery}.

Our experimental results indicate that \methodname{} allows to train a model that is up to $3.35\times$ smaller than other methods, with only marginal accuracy loss. Compared to \textsc{PowerSGD}, \textsc{Signum}, and vanilla SGD, \methodname{} achieves $1.22\times$, $1.52\times$, and $1.74\times$ end-to-end speedups respectively for ResNet-18 trained on CIFAR-10 while reaching to the same accuracy as vanilla SGD. \methodname{} leads to a model with $1.3M$ fewer parameters while reaching $1.76\%$ higher top-1 test accuracy than EB Train on the ImageNet dataset. Compared to LTH, \methodname{} leads to $5.67\times$ end-to-end speedup for achieving the same model compression ratio for VGG-19 on CIFAR-10. We also demonstrate that the performance of \methodname{} is stable under the ``mixed-precision training" implemented by PyTorch AMP. 
Our code is publicly available for reproducing our results\footnote{\url{https://github.com/hwang595/Pufferfish}}.
\vspace{-2 mm}
\subsection{Experimental setup and implementation details}
\paragraph{Setup.} \methodname{} is implemented in PyTorch~\cite{paszke2019pytorch}. We experiment using two implementations. The first implementation we consider is a data-parallel model training API, \ie DDP in PyTorch. However, as the gradient computation and communication are overlapped in DDP\footnote{the computed gradients are buffered and communicated immediately when hitting a certain buffer size, \eg 25MB.}, it is challenging to conduct a breakdown runtime analysis in DDP.  We thus also come up with a prototype \texttt{allreduce}-based distributed implementation that decouples the computation and communication to benchmark the breakdown runtime of \methodname{} and other baselines. Our prototype distributed implementation is based on \texttt{allreduce} in PyTorch and the NCCL backend. 
All our experiments are deployed on a distributed cluster consisting of up to 16 \texttt{p3.2xlarge} (Tesla V100 GPU equipped) instances on Amazon EC2.
\vspace{-2 mm}
\paragraph{Models and Datasets.}
The datasets considered in our experiments are CIFAR-10~\cite{krizhevsky2009learning}, ImageNet (ILSVRC2012)~\cite{deng2009imagenet}, the WikiText-2 datasets~\cite{merity2016pointer}, and the WMT 2016 German-English translation task data~\cite{elliott2016multi30k}. For the image classification tasks on CIFAR-10, we considered VGG-19-BN (which we refer to as VGG-19) \citep{simonyan2014very} and ResNet-18~\cite{he2016deep}. For ImageNet,  we run experiments with ResNet-50 and WideResNet-50-2 \citep{zagoruyko2016wide}. For the WikiText-2 dataset, we considered a 2-layer stacked LSTM model. For the language translation task, we consider a $6$-layer Transformer architecture \citep{vaswani2017attention}. More details about the datasets and models can be found in the Appendix.
\vspace{-2 mm}
\paragraph{Implementation details and optimizations.} In our prototype distributed implementation, the \texttt{allreduce} operation starts right after all compute nodes finish computing the gradient. 
An important implementation-level optimization we conduct is that we pack all gradient tensors into one flat buffer, and only call the \texttt{allreduce} operation \textbf{once} per iteration. The motivation for such an optimization is that \methodname{} factorizes the full-rank layer $W_l$ to two smaller layers, \ie $U_l, V^\top_l$. Though the communication cost of the \texttt{allreduce} on each smaller layer is reduced, the total number of \texttt{allreduce} calls is doubled (typically an \texttt{allreduce} is required per layer to synchronize the gradients across the distributed cluster). According to the run-time cost model of the ring-allreduce~\citep{thakur2005optimization}, 
each \texttt{allreduce} call introduces a network latency proportional to the product of the number of compute nodes and average network latency.
This is not a negligible cost. Our optimization strategy aims at minimizing the additional latency overhead and leads to good performance improvement based on our tests.  
For a fair comparison, we conduct the same communication optimization for all considered baselines. 

\begin{table}[ht]
    \vspace{-2mm}
	\caption{The results (averaged across $3$ independent trials with different random seeds) of \methodname{} and the vanilla 2-layer stacked LSTMs trained over the WikiText-2 dataset (since the embedding layer is just a look up table, we do not count it when calculating the MACs).}
	\vspace{-1 mm}
	\label{table:lstm-main-results}
	\begin{center}
      \scriptsize{
		\begin{tabular}{ccc}
		\toprule \textbf{Model archs.}
		& Vanilla LSTM & \methodname{} LSTM 
		\bigstrut\\
		\midrule
		\# Params. & $85,962,278$ & $67,962,278$ \bigstrut\\
		Train Ppl. & $52.87 \pm 2.43$ & $62.2\pm 0.74$\bigstrut\\
		Val Ppl. & $92.49\pm 0.41$ & $93.62 \pm 0.36$\bigstrut\\
		Test Ppl. & $88.16\pm 0.39$ & $88.72 \pm 0.24$\bigstrut\\
		MACs & $18$M & $9$M \bigstrut\\
		\bottomrule
		\end{tabular}}%
    \vspace{-6mm}
	\end{center}
\end{table}

\begin{table}[ht]
	\caption{The results (averaged across $3$ independent trials with different random seeds) of \methodname{} and vanilla 6-layer Transformers trained over the WMT 2016 German to English Translation Task.}
	\vspace{-1 mm}
	\label{table:transformer-main-results}
	\begin{center}
      \scriptsize{
		\begin{tabular}{ccc}
		\toprule \textbf{Model archs.}
		& Vanilla Transformer & \methodname{} Transformer 
		\bigstrut\\
		\midrule
		\# Params. & $48,978,432$ & $26,696,192$ \bigstrut\\
		Train Ppl . & $13.68\pm 0.96$ & $\bf{10.27 \pm 0.65}$ \bigstrut\\
		Val. Ppl . & $11.88\pm 0.43$ & $\bf{7.34 \pm 0.12}$ \bigstrut\\
		Val. BLEU & $19.05\pm 0.59$ & $\bf{26.87\pm 0.17}$ \bigstrut\\
		\bottomrule
		\end{tabular}}%
	\vspace{-5mm}
	\end{center}
\end{table}

\begin{table}[ht]
	\caption{The results (averaged across $3$ independent trials with different random seeds) of \methodname{} and vanilla VGG-19 and ResNet-18 trained over the CIFAR-10 dataset. Both full-precision training (FP32) and ``mixed-precision training" (AMP) results are reported.}
	\vspace{-2 mm}
	\label{table:cifar10-main-results}
	\begin{center}
      \scriptsize{
		\begin{tabular}{cccc}
		\toprule \textbf{Model Archs.}
		& \# Params. & Test Acc. (\%) & MACs (G) 
		\bigstrut\\
		\midrule
		Vanilla VGG-19 (FP32) & $20,560,330$ & $93.91 \pm 0.01$ & $0.4$ \bigstrut\\
		\methodname{} VGG-19 (FP32) & $8,370,634$ & $93.89\pm 0.14$ & $0.29$ \bigstrut\\
		Vanilla VGG-19 (AMP) & $20,560,330$ & $94.12\pm 0.08$ & N/A \bigstrut\\
		\methodname{} VGG-19 (AMP) & $8,370,634$ & $93.98\pm 0.06$ & N/A \bigstrut\\
		Vanilla ResNet-18 (FP32) & $11,173,834$ & $95.09\pm 0.01$ & $0.56$ \bigstrut\\
		\methodname{} ResNet-18 (FP32) & $3,336,138$ & $94.87\pm 0.21$ & $0.22$ \bigstrut\\
		Vanilla ResNet-18 (AMP) & $11,173,834$ & $95.02\pm 0.1$ & N/A \bigstrut\\
		\methodname{} ResNet-18 (AMP) & $3,336,138$ & $94.70\pm 0.37$ & N/A \bigstrut\\
		\bottomrule
		\end{tabular}}%
    \vspace{-4mm}
	\end{center}
\end{table}

\begin{table*}[ht]
	\caption{The results of the vanilla and \methodname{} ResNet-50 and WideResNet-50-2 models trained on the ImageNet dataset. For the ResNet-50 results, both full precision training (FP32) and mixed-precision training (AMP) are provided. For the AMP training, MACs are not calculated.}
	\vspace{-1 mm}
	\label{table:imagenet-main-results}
	\begin{center}
		 \scriptsize{
		\begin{tabular}{ccccc}
		\toprule \textbf{Model Archs.}
		& Number of Parameters &  Final Test Acc. (Top-1) & Final Test Acc. (Top-5) & MACs (G) 
		\bigstrut\\
		\midrule
		Vanilla WideResNet-50-2 (FP32) & $68,883,240$ & $78.09\%$ & $94.00\%$ & $11.44$ \bigstrut\\
		\methodname{} WideResNet-50-2 (FP32) & $40,047,400$ & $77.84\%$ & $93.88\%$ & $9.99$ \bigstrut\\
		Vanilla ResNet-50 (FP32) & $25,557,032$ & $76.93\%$ & $93.41\%$ & $4.12$ \bigstrut\\
		\methodname{} ResNet-50 (FP32) & $15,202,344$ & $76.43\%$ & $93.10\%$ & $3.6$ \bigstrut\\
		Vanilla ResNet-50 (AMP) & $25,557,032$ & $76.97\%$ & $93.35\%$ & N/A \bigstrut\\
		\methodname{} ResNet-50 (AMP) & $15,202,344$ & $76.35\%$ & $93.22\%$ & N/A \bigstrut\\
		\bottomrule
		\end{tabular}}%
    \vspace{-6mm}
	\end{center}
\end{table*}

\begin{table}[ht]
    \vspace{-1 mm}
	\caption{The runtime mini-benckmark results of \methodname{} and vanilla VGG-19 and ResNet-18 networks training on the CIFAR-10 dataset. Experiment running on a single V100 GPU with batch size at $128$, results averaged over $10$ epochs; under the reproducible cuDNN setup with \texttt{cudnn.benckmark} disabled and \texttt{cudnn.deterministic} enabled; Speedup calculated based on the averaged runtime.}
	\vspace{-1 mm}
	\label{table:mini-benchmark}
	\begin{center}
      \scriptsize{
		\begin{tabular}{cccc}
		\toprule \textbf{Model Archs.}
		&  Epoch Time (sec.) & Speedup & MACs (G) 
		\bigstrut\\
		\midrule
		Vanilla VGG-19 & $13.51 \pm 0.02$ & $-$ & $0.4$ \bigstrut\\
		\methodname{} VGG-19  & $\bf{11.02\pm 0.01}$ & $\bf{1.23\times}$& $\bf{0.29}$ \bigstrut\\
		Vanilla ResNet-18  & $18.89\pm 0.07$ & $-$ & $0.56$ \bigstrut\\
		\methodname{} ResNet-18 & $\bf{12.78\pm 0.03}$ & $\bf{1.48\times}$ & $\bf{0.22}$ \bigstrut\\
		\bottomrule
		\end{tabular}}%
     \vspace{-3mm}
	\end{center}
\end{table}


\vspace{-2 mm}
\paragraph{Hyper-parameters for \methodname{}.} 

For all considered model architectures, we use a global rank ratio of $0.25$, \eg for a convolution layer with an initial rank of $64$, \methodname{} sets $r = 64\times 0.25=16$. For the LSTM on WikiText-2 experiment, we only factorize the LSTM layers and leave the tied embedding layer as is. Allocating the optimal rank for each layer can lead to better final model accuracy and smaller model sizes as discussed in~\cite{idelbayev2020low}. However, the search space for the rank allocation problem is large. One potential way to solve that problem is to borrow ideas from the literature of neural architectural search (NAS), which we leave as future work. We tune the initial low-rank layer index, \ie $K$ and the vanilla warm-up training period to balance the hybrid model size and the final model accuracy. More details of the hyper-parameters of \methodname{} can be found in the Appendix. 
\vspace{-2 mm}
\subsection{Results}
\begin{figure*}[t]
	\centering
	\includegraphics[width=0.25\textwidth]{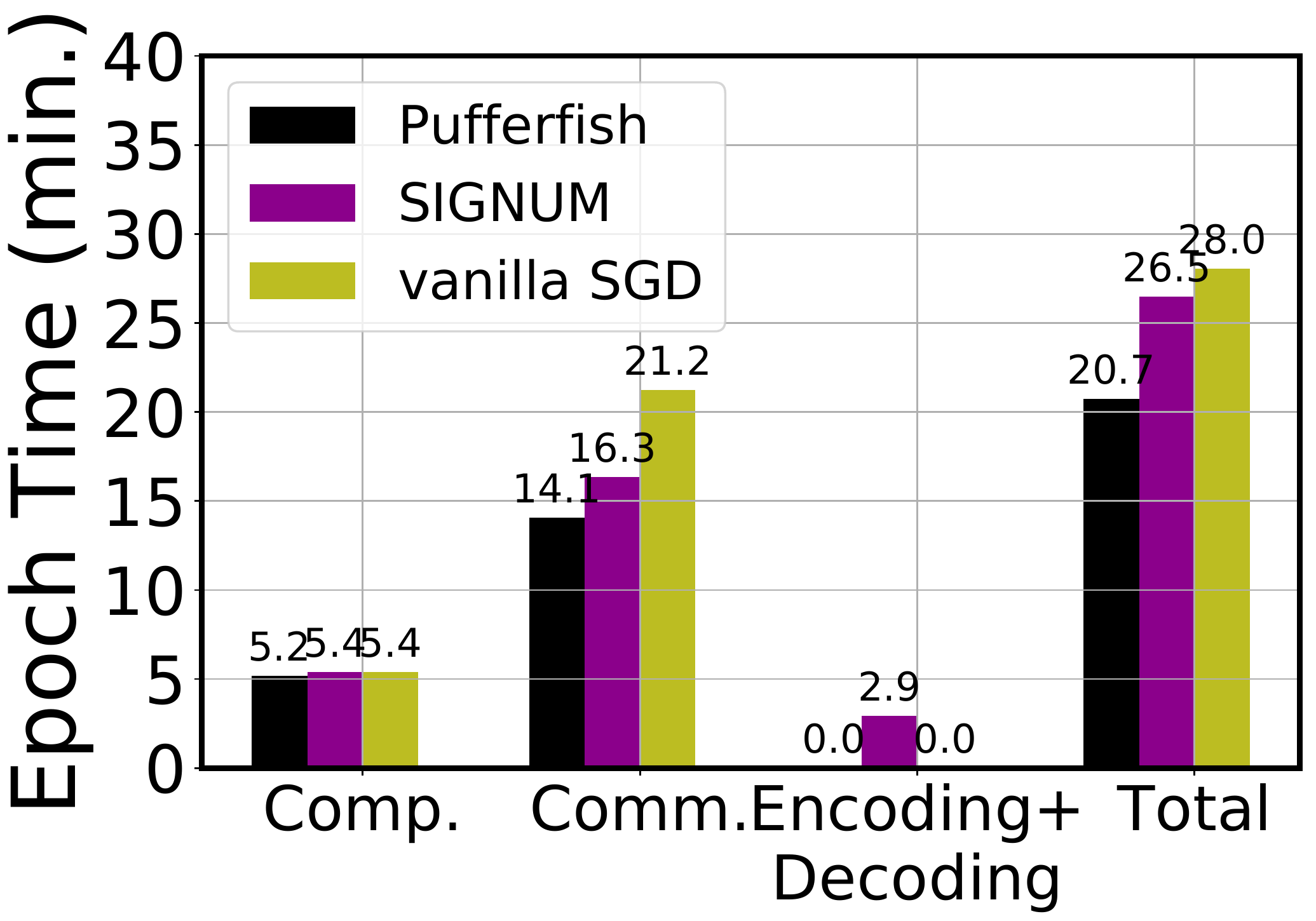}
    \includegraphics[width=0.25\textwidth]{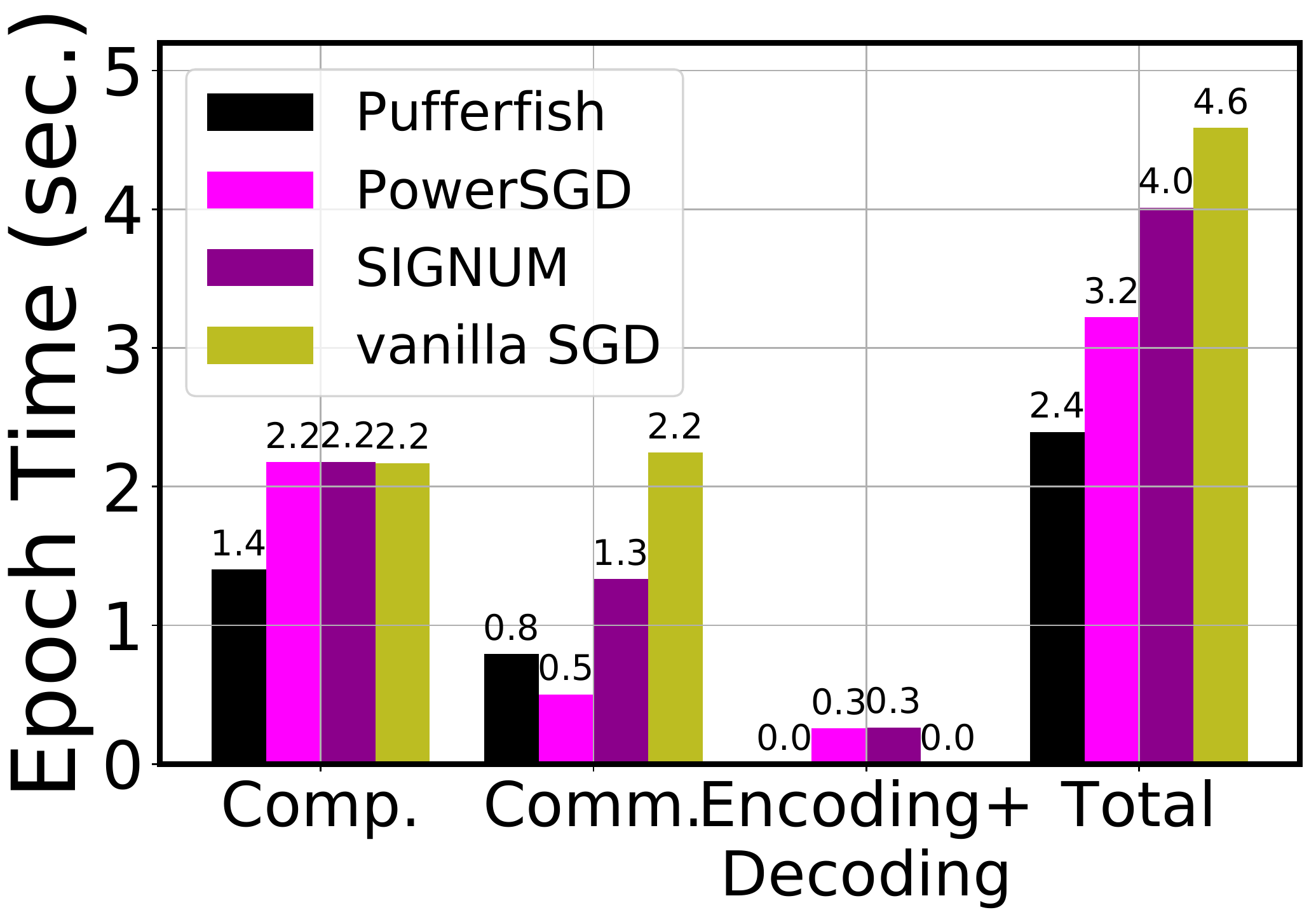}
	\includegraphics[width=0.25\textwidth]{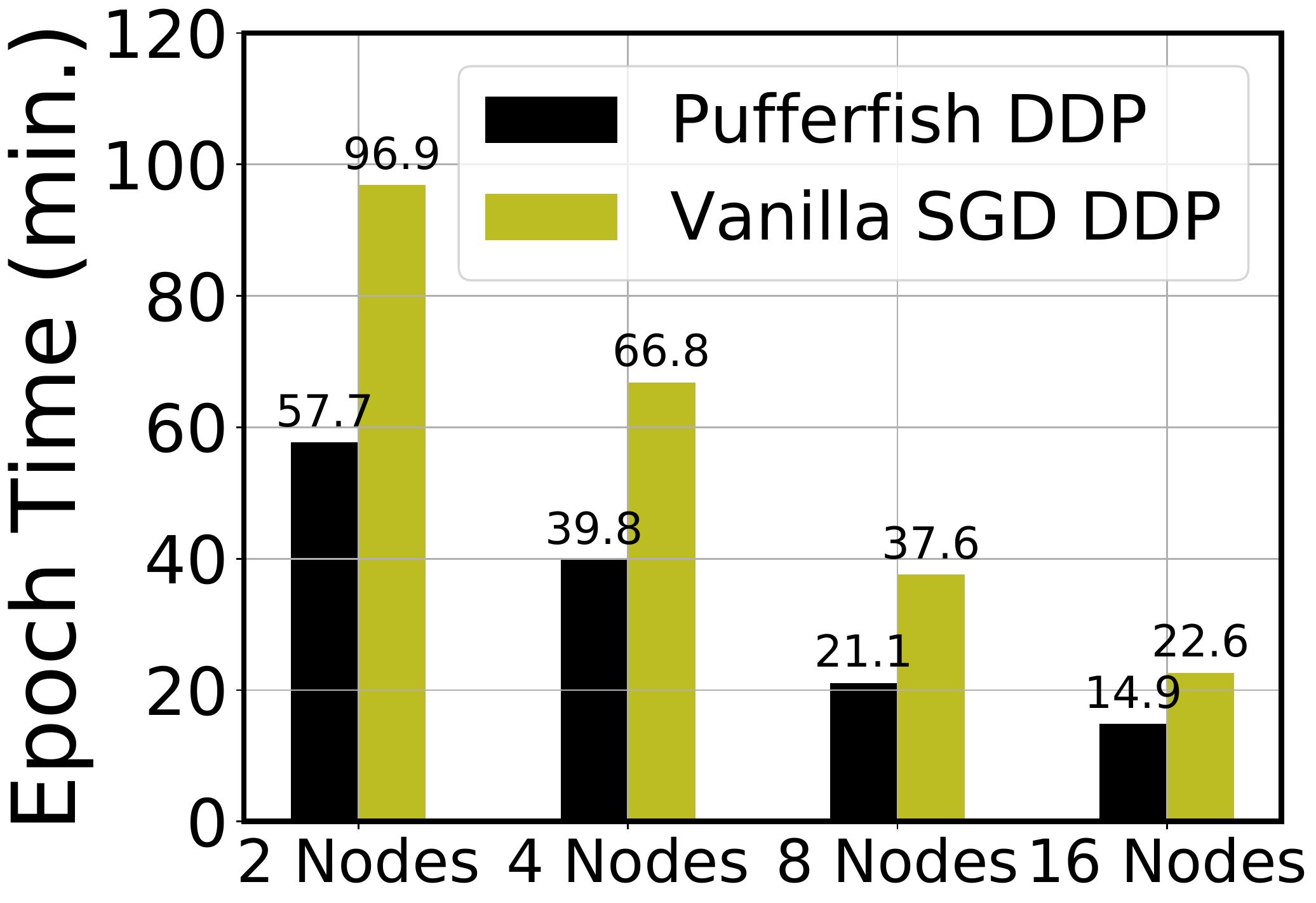}\\
	\vspace{-2mm}
	\subfigure[Proto. ResNet-50, ImageNet]{\includegraphics[width=0.25\textwidth]{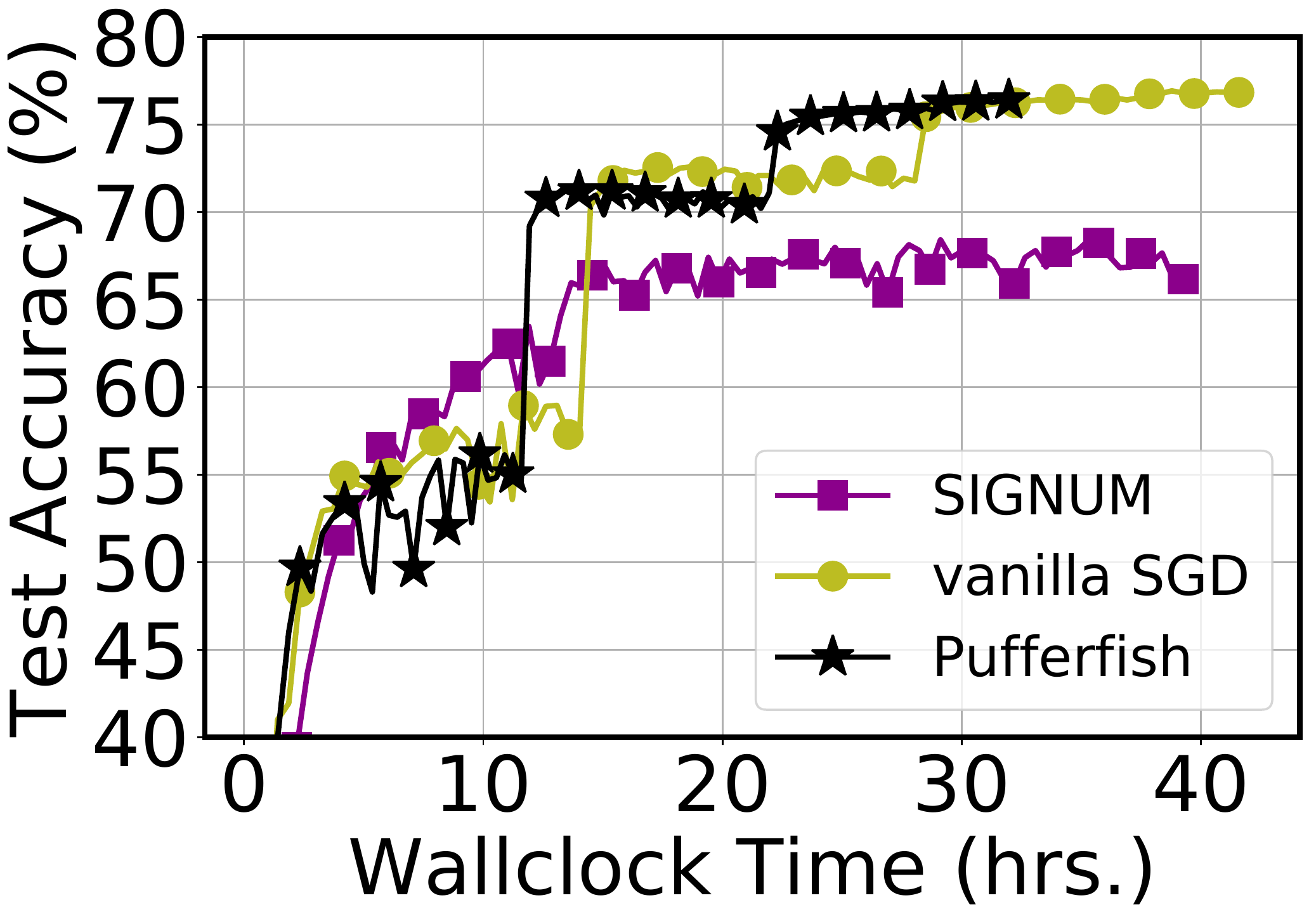}\label{fig:proto-imagenet-resnet50}}
	\subfigure[Proto. ResNet-18, CIFAR-10]{\includegraphics[width=0.25\textwidth]{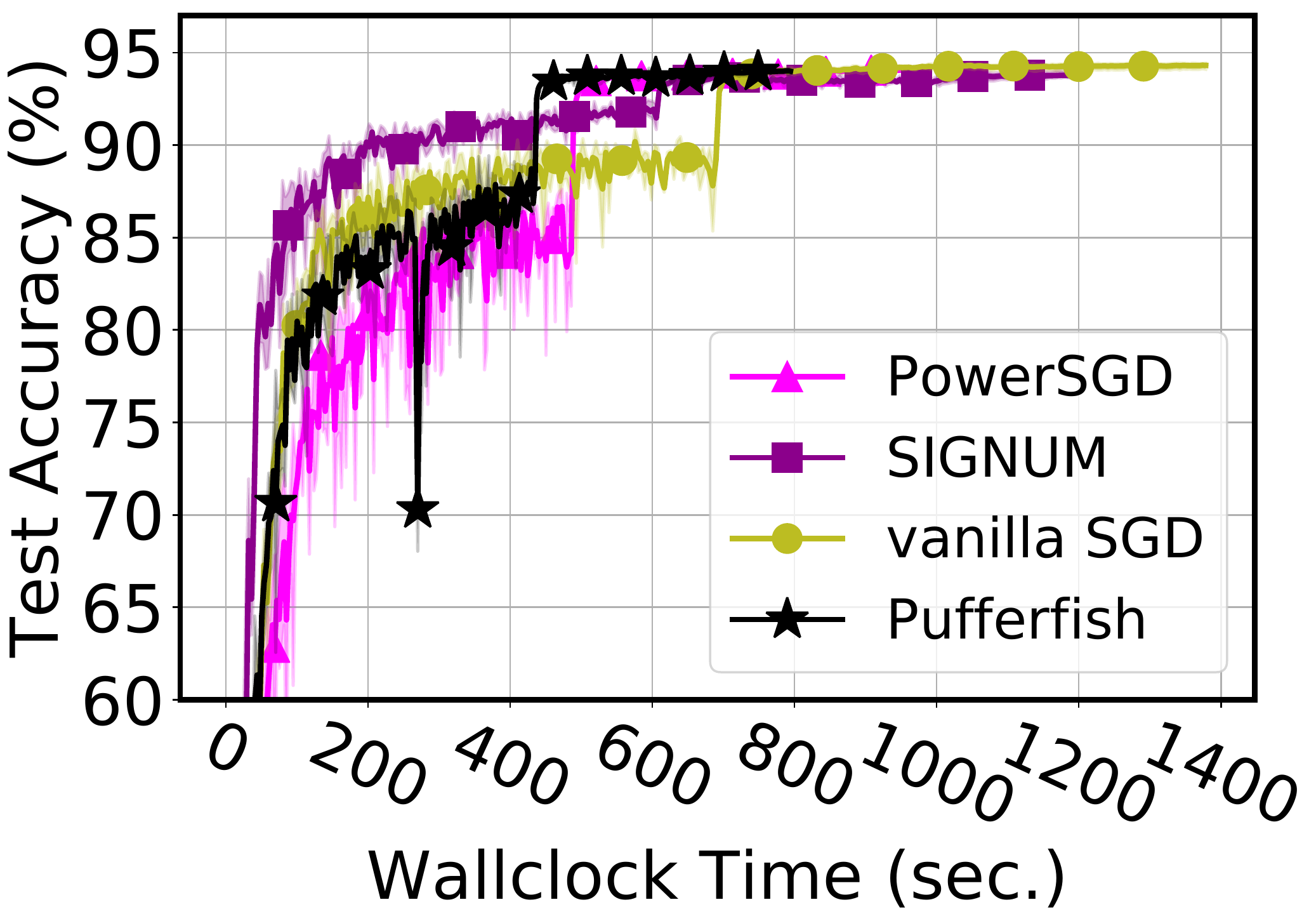}\label{fig:proto-cifar10-resnet18}}
	\subfigure[DDP ResNet-50, ImageNet]{\includegraphics[width=0.25\textwidth]{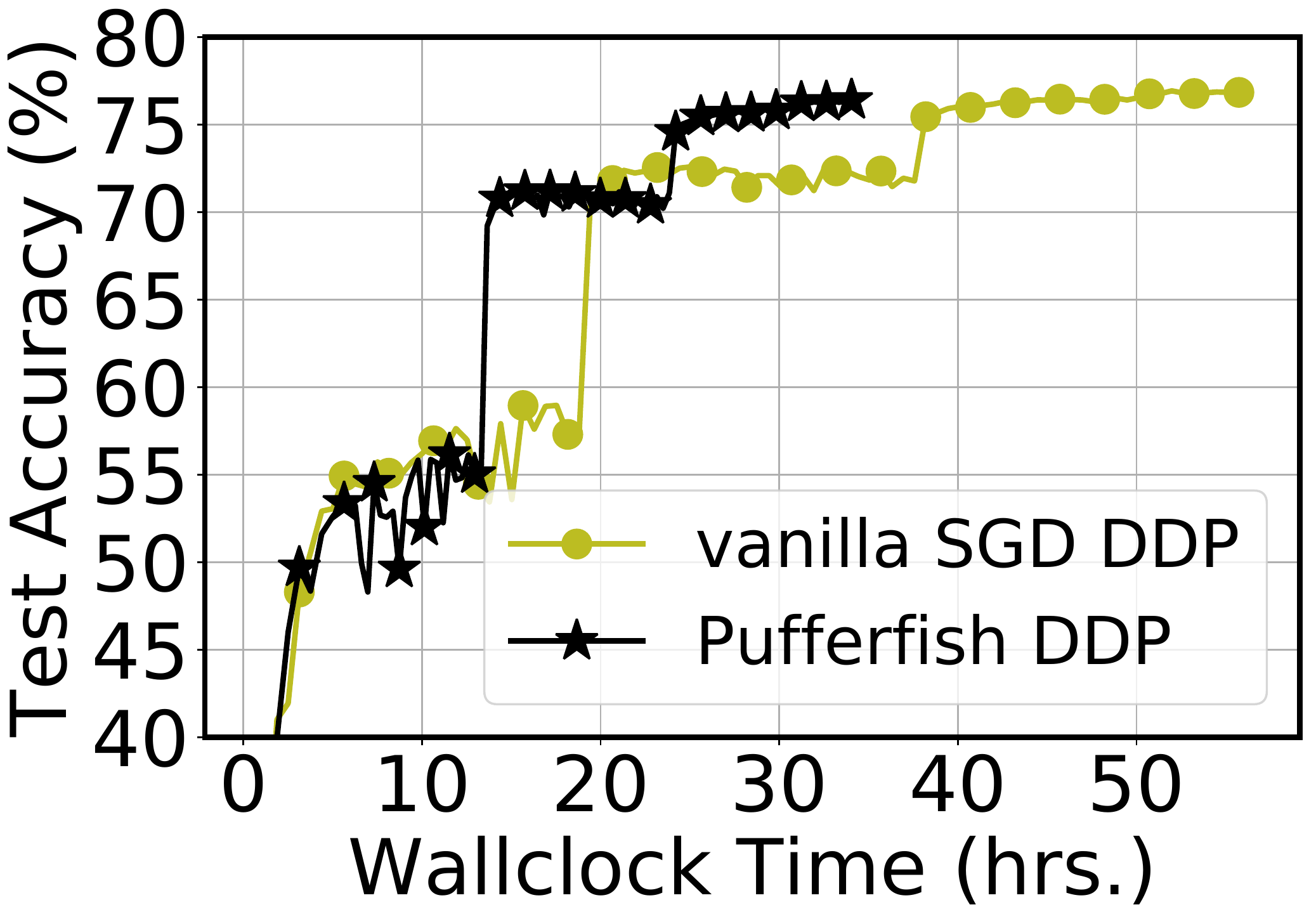}\label{fig:ddp-resnet50-ddp}}	
	\vspace{-4 mm}
	\caption{(a) Breakdown per-epoch runtime analysis (top) and end-to-end convergence (bottom) results for vanilla SGD, \methodname{}, and \textsc{signum} over ResNet-50 trained on the ImageNet dataset. Where \texttt{Comm.} and \texttt{Comp.} stands for computation and communication costs; (b) Breakdown per-epoch runtime analysis (top) and end-to-end convergence (bottom) results for vanilla SGD, \methodname{}, \textsc{signum}, and PowerSGD over ResNet-18 trained on CIFAR-10; (c) The scalability of \methodname{} compared to vanilla SGD for ResNet-50 training on ImageNet using PyTorch DDP over the distributed clusters that consist of $2, 4, 8, 16$ nodes (top); End-to-end convergence for vanilla SGD and \methodname{} with PyTorch DDP under the cluster with $8$ nodes (bottom).
	}
	\vspace{-4 mm}
\end{figure*}

\paragraph{Parameter reduction and model accuracy.}  We extensively study the effectiveness of \methodname{}, and the comprehensive numerical results are shown in Table~\ref{table:lstm-main-results}, \ref{table:transformer-main-results}, \ref{table:cifar10-main-results}, and \ref{table:imagenet-main-results}. The main observation is that \methodname{} effectively reduces the number of parameters while introducing only marginal accuracy loss. In particular, \methodname{} ResNet-18 is $3.35\times$ smaller than vanilla ResNet-18 with only $0.22\%$ accuracy loss. Surprisingly, the \methodname{} Transformer leads to even better validation perplexity and test BLEU scores than the vanilla Transformer. One potential reason behind that is that factorizing the Transformer introduces some implicit regularization, leading to better generalization. Apart from the full precision training over FP32, we also conduct mixed-precision experiments over PyTorch AMP on both CIFAR-10 and ImageNet. Our results generally demonstrate that the performance of \methodname{} remains stable under mixed-precision training. We measure the computational complexity using ``\textit{multiply–accumulate operations}" (MACs) \footnote{\url{https://en.wikipedia.org/wiki/Multiply\%E2\%80\%93accumulate_operation}}. The MAC results are shown in Table~\ref{table:lstm-main-results}, \ref{table:cifar10-main-results}, and \ref{table:imagenet-main-results}. The computation complexity is estimated by passing a single input through the entire network, \eg for the CIFAR-10 dataset, we simulate a color image with size $32\times 32 \times 3$ and pass it to the networks. For the LSTM network, we assume a single input token is with batch size at $1$. We only report the MACs of forward pass. 
\methodname{} reduces the MACs of the vanilla model to up to $2.55\times$ over ResNet-18 on CIFAR-10.

\vspace{-1mm}
\paragraph{Runtime mini-benchmark.} It is of interest to investigate the actual speedup of the factorized networks as they are dense and compact. We thus provide mini-benchmark runtime results over VGG-19 and ResNet-18 on the CIFAR-10 dataset. We measure the per-epoch training speed of the factorized networks used in \methodname{} and the vanilla networks on a single V100 GPU with batch size at $128$. The results are shown in Table~\ref{table:mini-benchmark}. We report the results (averaged over $10$ epochs) under the reproducibility optimized cuDNN environment, \ie \texttt{cudnn.benckmark} disabled and \texttt{cudnn.deterministic} enabled. The results indicate that the factorized networks enjoy promising runtime speedups, \ie $1.23\times$ and $1.48\times$ over the vanilla VGG-19 and ResNet-18 respectively. We also study the runtime of the factorized networks under the speed optimized cuDNN setting, \ie \texttt{cudnn.benckmark} enabled and \texttt{cudnn.deterministic} disabled, the results can be found in the Appendix. 
\vspace{-2mm}
\paragraph{Computation and communication efficiency.} To benchmark the computation and communication costs of \methodname{} under a distributed environment, we conduct a per-epoch breakdown runtime analysis and compare it to vailla SGD and \textsc{Signum} on ResNet-50, trained over ImageNet. The experiment is conducted over $16$ \texttt{p3.2xlarge} EC2 instances. We set the global batch size at $256$ ($16$ per node). We use tuned hyper-parameters for all considered baselines. The result is shown in Figure~\ref{fig:proto-imagenet-resnet50} where we observe that the \methodname{} ResNet-50 achieves $1.35\times$ and $1.28\times$ per-epoch speedups compared to vanilla SGD and \textsc{Signum} respectively. Note that though \textsc{Signum} achieves high compression ratio, it is not compatible with \texttt{allreduce}, thus \texttt{allgather} is used instead in our \textsc{Signum} implementation. However, \texttt{allgather} is less efficient than \texttt{allreduce}, which hurts the communication efficiency of \textsc{signum}. The effect has also been observed in the previous literature~\citep{vogels2019powersgd}. We extend the per-epoch breakdown runtime analysis to ResNet-18 trainining on CIFAR-10 where we compare \methodname{} to \textsc{PowerSGD}, \textsc{signum}, and vanilla SGD. The experiments are conducted over $8$ \texttt{p3.2xlarge} EC2 instances with the global batch size at $2048$  ($256$ per node). We use a linear learning rate warm-up for $5$ epochs from $0.1$ to $1.6$, which follows the setting in~\cite{vogels2019powersgd,goyal2017accurate}. For \textsc{PowerSGD}, we set the rank at $2$, as it matches the same accuracy compared to vanilla SGD~\cite{vogels2019powersgd}. The results are shown in Figure~\ref{fig:proto-cifar10-resnet18}, from which we observe that \methodname{} achieves $1.33\times, 1.67\times, 1.92\times$ per-epoch speedups over \textsc{PowerSGD}, \textsc{signum}, and vanilla SGD respectively. Note that  \methodname{} is slower than \textsc{PowerSGD} in the communication stage since \textsc{PowerSGD} massively compresses gradient and is also compatible with \texttt{allreduce}. However, \methodname{} is faster for gradient computing and bypasses the gradient encoding and decoding steps. Thus, the overall epoch time cost of \methodname{} is faster than \textsc{PowerSGD}. Other model training overheads, \eg data loading and pre-processing, gradient flattening, and etc are not included in the ``computation" stage but in the overall per-epoch time.
Since \methodname{} only requires to modify the model architectures instead of gradients, it is directly compatible with current data-parallel training APIs, \eg DDP in PyTorch. Other gradient compression methods achieve high compression ratio, but they are not directly compatible with DDP without significant engineering effort. 
For PyTorch DDP, we study the speedup of \methodname{} over vanilla distributed training, measuring the per-epoch runtime on ResNet-50 and ImageNet over distributed clusters of size $2, 4, 8,$ and $16$. We fix the per-node batch size at $32$ following the setup in ~\cite{goyal2017accurate}. The results are shown in Figure \ref{fig:ddp-resnet50-ddp}. We observe that \methodname{} consistently outperforms vanilla ResNet-50. In particular, on the cluster with  $16$ nodes, \methodname{} achieves $1.52\times$ per epoch speedup.
\vspace{-2 mm}
\paragraph{End-to-end speedup.} We study the end-to-end speedup of \methodname{} against other baselines under both our prototype implementation and PyTorch DDP. The experimental setups for the end-to-end experiment are identical to our per-epoch breakdown runtime analysis setups. All reported runtimes include the overheads of the SVD factorization and vanilla warm-up training. The ResNet-50 on  ImageNet  convergence results with our prototype implementation are shown in Figure~\ref{fig:proto-imagenet-resnet50}. We observe that to finish the entire $90$ training epochs, \methodname{} attains $1.3\times$ and $1.23\times$ end-to-end speedups compared to vanilla SGD and \textsc{Signum} respectively. The ResNet-18 on CIFAR-10 convergence results are shown in Figure~\ref{fig:proto-cifar10-resnet18}. For faster vanilla warm-up training in \methodname{}, we deploy \textsc{PowerSGD} to compress the gradients. We observe that it is generally better to use a slightly higher rank for \textsc{PowerSGD} in the vanilla warm-up training period of \methodname{}. In our experiments, we use \textsc{PowerSGD} with rank $4$ to warm up \methodname{}. We observe that to finish the entire $300$ training epochs, \methodname{} attains $1.74\times, 1.52\times, 1.22\times$ end-to-end speedup compared to vanilla SGD, \textsc{signum}, and \textsc{PowerSGD} respectively. \methodname{} reaches to the same accuracy compared to vanilla SGD. Moreover, we extend the end-to-end speedup study under PyTorch DDP where we compare \methodname{} with vanilla SGD under $8$ EC2 \texttt{p3.2xlarge} instances. The global batch size is $256$ ($32$ per node). The results are shown in Figure~\ref{fig:ddp-resnet50-ddp} where we observe that to train the model for $90$ epochs, \methodname{} achieves $1.64\times$ end-to-end speedup compared to vanilla SGD. We do not study the performance of \textsc{signum} and \textsc{PowerSGD} under DDP since they are not directly compatible with DDP.
\begin{table*}[ht]
	\caption{Comparison of Hybrid ResNet-50 model compared to the Early-Bird Ticket structure pruned (EB Train) ResNet-50 model results with prune ratio $pr$ at $30\%, 50\%, 70\%$ over the ImageNet dataset}
	\label{table:comparison-eb-train}
	\begin{center}
		 \scriptsize{
		\begin{tabular}{ccccc}
		\toprule \textbf{Model architectures}
		& \# Parameters &  Final Test Acc. (Top-1) & Final Test Acc. (Top-5) & MACs (G) 
		\bigstrut\\
		\midrule
		vanilla ResNet-50 & $25,610,205$ & $75.99\%$ & $92.98\%$ & $4.12$ \bigstrut\\
		\methodname{} ResNet-50 & $15,202,344$ & $75.62\%$ & $92.55\%$ & $3.6$ \bigstrut\\
		EB Train ($pr=30\%$) & $16,466,787$ & $73.86\%$ & $91.52\%$ & $2.8$ \bigstrut\\
		EB Train ($pr=50\%$) & $15,081,947$ & $73.35\%$ & $91.36\%$ & $2.37$ \bigstrut\\
		EB Train ($pr=70\%$) & $7,882,503$ & $70.16\%$ & $89.55\%$ & $1.03$ \bigstrut\\
		\bottomrule
		\end{tabular}}%
 	\vspace{-8mm}
	\end{center}
\end{table*}
\vspace{-4 mm}
\paragraph{Comparison with structured pruning.}
We compare \methodname{} with the EB Train method where structured pruning is conducted over the channel dimensions based on the activation values during the early training phase~\citep{you2019drawing}. EB Train finds compact and dense models. The result is shown in Table \ref{table:comparison-eb-train}. We observe that compared to EB Train with prune ratio $(pr)=30\%$, \methodname{} returns a model with $1.3M$ fewer parameters while reaching $1.76\%$ higher top-1 test accuracy. The EB Train experimental results are taken directly from the original paper~\citep{you2019drawing}. To make a fair comparison, we train \methodname{} with the same hyper-parameters that EB Train uses, \eg removing label smoothing and only decaying the learning rate at the $30$-th and the $60$-th epochs with the factor $0.1$.
\vspace{-2 mm}
\begin{figure}[ht]
	\centering
	\subfigure[Model size \textit{vs} Runtime]{\includegraphics[width=0.225\textwidth]{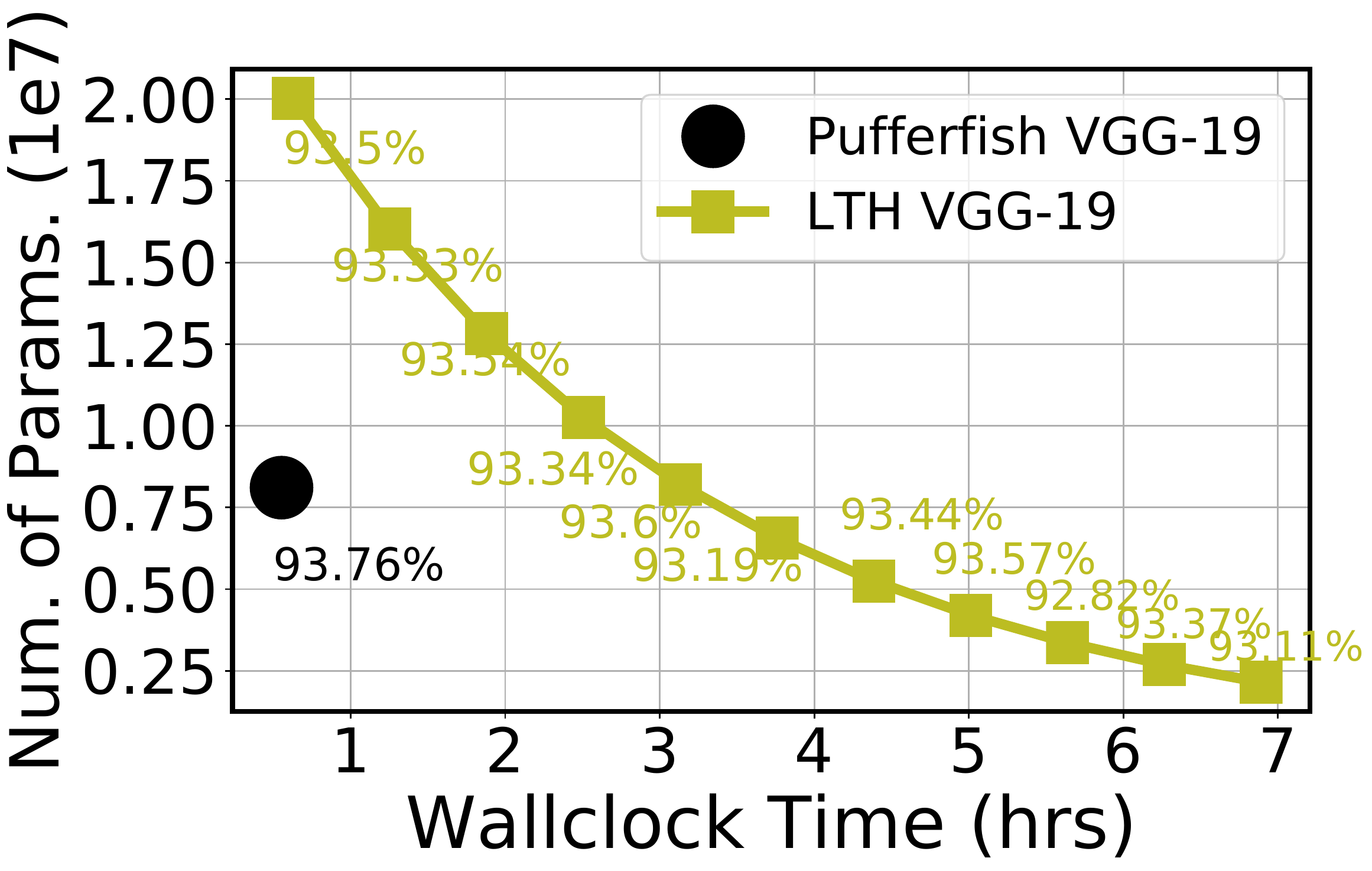}\label{fig:lth-comp-time}}
	\subfigure[Model size \textit{vs} Test Acc.]{\includegraphics[width=0.225\textwidth]{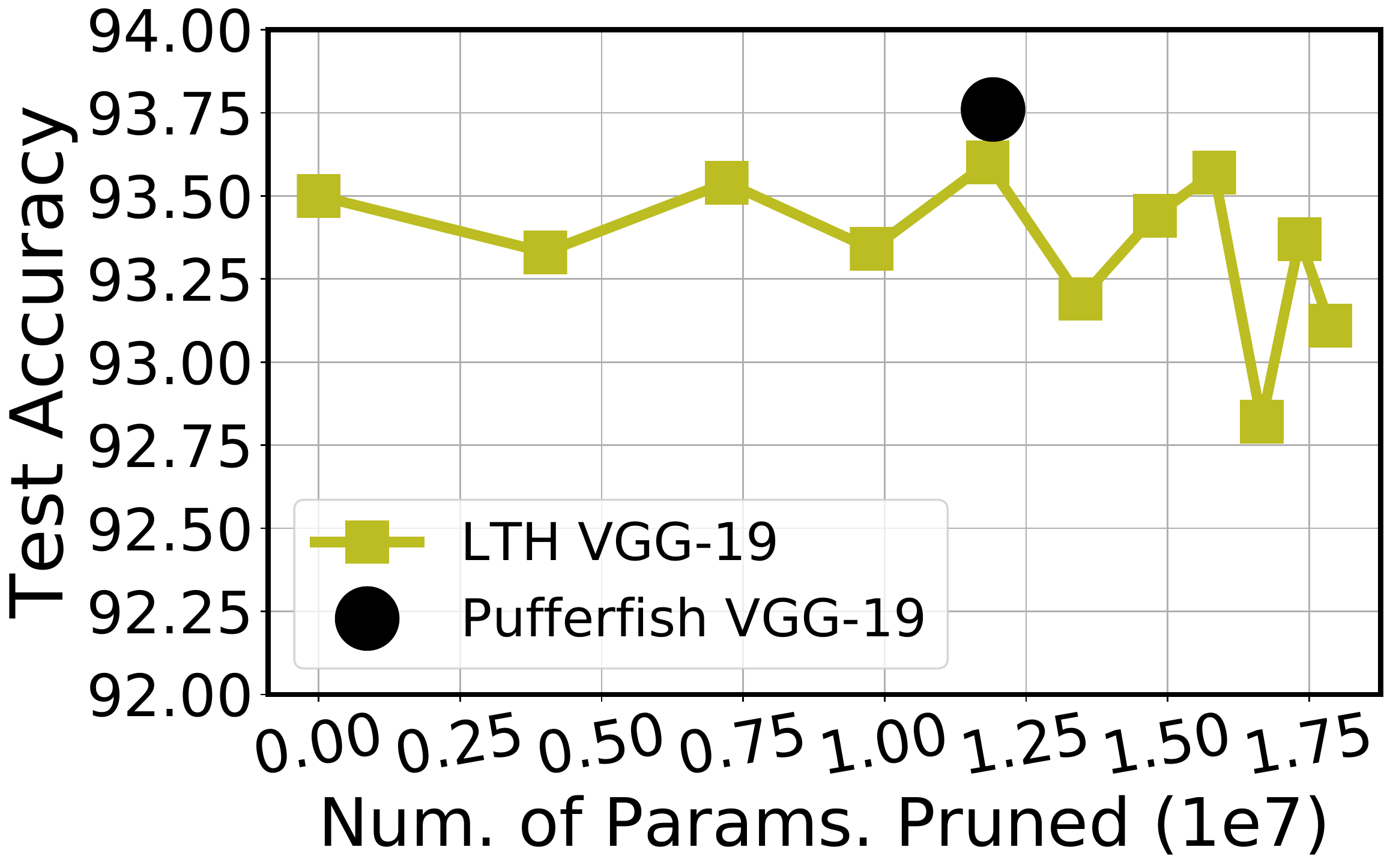}\label{fig:lth-comp-acc}}
	\vspace{-4 mm}
	\caption{The performance comparison between \methodname{} and LTH over a VGG-19 model trained over the CIFAR-10 dataset: (a) the number of parameters \textit{v.s.} wall-clock runtime; (b) the number of parameters pruned \textit{v.s.} the test accuracy.
	}
\vspace{-6mm}
\end{figure}
\vspace{-2mm}
\paragraph{Comparison with LTH.}
The recent LTH literature initiated by Frankle et al.~\cite{frankle2018lottery}, indicates that dense, randomly-initialized networks contain sparse subnetworks (referred to as ``\textit{winning tickets}") that---when trained in isolation---reach test accuracy comparable to the original network~\citep{frankle2018lottery}. To find the winning tickets, an iterative pruning algorithm is conducted, which trains, prunes, and rewinds the remaining unpruned elements to their original random values repeatedly. Though LTH can compress the model massively without significant accuracy loss, the iterative pruning is computationally heavy. We compare \methodname{} to LTH across model sizes and computational costs on VGG-19 trained with CIFAR-10. The results are shown in Figure \ref{fig:lth-comp-time}, \ref{fig:lth-comp-acc} where we observe that to prune the same number of parameters, LTH costs $5.67\times$ more time than \methodname{}.
\vspace{-4mm}
\paragraph{Ablation study.}
We conduct an ablation study on the accuracy loss mitigation methods in \methodname{}, \ie hybrid network and vanilla warm-up training. The results on ResNet-18+CIFAR-10 and LSTM+WikiText-2 are shown in Table~\ref{table:ablation-resnet18-cifar10} and Table~\ref{table:ablation-lstm}, which indicate that the hybrid network and vanilla warm-up training methods help to mitigate the accuracy loss effectively. Results on the other datasets can be found in the Appendix.

\begin{table}[ht]
	\caption{The effect of vanilla warm-up training and hybrid network architectures of \methodname{} of the low rank ResNet-18 trained over the CIFAR-10 dataset. Results are averaged across $3$ independent trials with different random seeds.}
	\vspace{-2 mm}
	\label{table:ablation-resnet18-cifar10}
	\begin{center}
		 \scriptsize{
		\begin{tabular}{ccc}
		\toprule \textbf{Methods}
		 & Test Loss & Test Acc. (\%) 
		\bigstrut\\
		\midrule
		Low-rank ResNet-18 & $0.31\pm 0.01$ & $93.75\pm 0.19$ \bigstrut\\
		Hybrid ResNet-18 (wo. vanilla warm-up) & $0.30\pm 0.02$ & $93.92\pm 0.45$ \bigstrut\\
		Hybrid ResNet-18 (w. vanilla warm-up)  & ${\bf 0.25}\pm 0.01$ & ${\bf94.87} \pm 0.21$ \bigstrut\\
		\bottomrule
		\end{tabular}%
		}
    \vspace{-4mm}
	\end{center}
\end{table}

\begin{table}[ht]
    \vspace{-3 mm}
	\caption{The effect of vanilla warm-up training on the low-rank LSTM trained over  WikiText-2. Results are averaged across $3$ independent trials with different random seeds.}
	\vspace{-1 mm}
	\label{table:ablation-lstm}
	\begin{center}
		 \scriptsize{
		\begin{tabular}{ccc}
		\toprule \textbf{Methods}
		 & Low-rank LSTM  & Low-rank LSTM \\
		 & (wo. vanilla warm-up) & (w. vanilla warm-up) \bigstrut\\
		\midrule
		 Train Ppl. & $68.04 \pm 2.98$ & $\bf{62.2}\pm 0.74$ \bigstrut\\
		 Val. Ppl. & $97.59 \pm 0.69$ & $\bf{93.62}\pm0.36$ \bigstrut\\
		 Test Ppl.  & $92.04 \pm 0.54$ & $\bf{88.72}\pm 0.24$ \bigstrut\\
		\bottomrule
		\end{tabular}%
		}
    \vspace{-8mm}
	\end{center}
\end{table}
\vspace{-4mm}
\paragraph{Limitations of \methodname{}.}
One limitation of \methodname{} is that it introduces two extra hyper-parameters, \ie the initial low-rank layer index $K$ and the vanilla warm-up epoch $E_{\text{wu}}$, hence hyperparameter tuning requires extra effort. Another limitation is that although \methodname{} reduces the parameters in ResNet-18 and VGG-19 models effectively for the CIFAR-10 dataset, it only finds $1.68\times$ and $1.72\times$ smaller models for ResNet-50 and WideResNet-50-2 in order to preserve good final model accuracy.
\vspace{-2mm}
\vspace{-2mm}
\section{Conclusion}
We propose \methodname{}, a communication and computation efficient distributed training framework. Instead of gradient compression, \methodname{} trains low-rank networks initialized by factorizing a partially trained full-rank model. The use of a hybrid low-rank model and warm-up training, allows \methodname{} to preserve the accuracy of the fully dense SGD trained  model, while effectively reducing its size. \methodname{} achieves high computation and communication efficiency and completely bypasses the gradient encoding and decoding, while yielding smaller and more accurate models compared pruning methods such the LTH and EB Train, while avoiding the burden of ``winning the lottery". 
\vspace{-2mm}

\paragraph{Acknowledgments}
This research is supported by an NSF CAREER Award \#1844951, two SONY Faculty Innovation Awards,  an AFOSR \& AFRL
Center of Excellence Award FA9550-18-1-0166, and an NSF TRIPODS Award \#1740707. The authors also thank Yoshiki Tanaka, Hisahiro Suganuma, Pongsakorn U-chupala, Yuji Nishimaki, and Tomoki Sato from SONY for invaluable discussions and feedback.
\bibliography{pufferfish}
\bibliographystyle{mlsys2021}

\appendix
\newpage
\section{Software versions used in the experiments}
Since we provide wall-clock time results, it is important to specify the versions of the libraries we used. For the full-precision (FP32) results, we used the \texttt{pytorch\_p36} virtual environment associated with the ``Deep Learning AMI (Ubuntu 18.04) Version 40.0 (ami-084f81625fbc98fa4)" on Amazon EC2, \ie \texttt{PyTorch 1.4.0} with $\text{CUDA} 10.1.243$. Since AMP is only supported after $1.6.0$ of PyTorch, we use \texttt{PyTorch 1.6.0} with CUDA 10.1.

\section{Detailed discussion on the computation complexity of various layers}
We provide the complexity and number of parameters in the vanilla and low-rank factorized layers returned by \methodname{} in Section \ref{sec:pufferfish} without providing any details. We give the detailed discussion here. 
\paragraph{FC layer.} We start from the FC layer, assuming the input $x\in \mathbb{R}^{m\times n}$, the computation complexity is simply $\mathcal{O}(mn)$ for $xW$ and $\mathcal{O}(mr + rn)$ for $(xU)V^\top$. For the number of parameters, the vanilla FC layer contains $mn$ parameters in total while the low-rank FC layer contains $r(m+n)$ parameters in total.

\paragraph{Convolution layer.} For a convolution layer, assuming the input is with dimension $x\in \mathbb{R}^{c_\text{in}\times H\times W}$ (the input ``image" has $c_{\text{in}}$ color channels and with size $H\times W$), the computation complexity of a vanilla convolution layer with weight $W \in \mathbb{R}^{c_{\text{in}}\times c_{\text{out}} \times k \times k}$ is $\mathcal{O}(c_{\text{in}}c_{\text{out}}k^2HW)$ for computing $W*x$ where $*$ is the linear convolution operation. And the low-rank factorized convolution layer with dimension $U \in \mathbb{R}^{c_{\text{in}}\times r \times k \times k}, V \in \mathbb{R}^{r\times c_{\text{out}} \times 1 \times 1}$ has the computation complexity at $\mathcal{O}(rc_{\text{in}}k^2HW)$ for $U*x$ and $\mathcal{O}(rHWc_{\text{out}})$ for convolving the output of $U*x$ with $V$. For the number of parameters, the vanilla convolution layer contains $c_{\text{in}}c_{\text{out}}k^2$ parameters in total while the low-rank convolution layer contains $c_{\text{in}}rk^2+rc_{\text{out}}$ parameters in total.

\paragraph{LSTM layer.} For the LSTM layer, the computation complexity is similar to the computation complexity of the FC layer. Assuming the tokenized input is with dimension $x\in \mathbb{R}^d$, and the concatenated input-hidden and hidden-hidden weights $W_i \in \mathbb{R}^{d\times 4h}, W_h \in \mathbb{R}^{4h \times h}$, thus the computation complexity of the forward propagation of a LSTM layer is $\mathcal{O}(4dh + 4h^2)$. And for the low-rank LSTM layer, the computation complexity becomes $\mathcal{O}(dr+4rh + 4hr+rh)$ (as mentioned in Section \ref{sec:pufferfish}, we assume that the same rank $r$ is used for both the input-hidden weight and hidden-hidden weight). For the number of parameters, the vanilla LSTM layer contains $4dh+4h^2$ parameters in total while the low-rank convolution layer contains $4(dr+rh) + 4(hr+rh)=4dr+12hr$ parameters in total.

\paragraph{Transformer.} For the encoder layer in the Transformer architecture, there are two main components, \ie the multi-head attention layer and the FFN layer. Note that, for the multi-head attention layer, the dimensions of the projection matrices are: The dimensions of the matrices are $Q, K, V \in \mathbb{R}^{n\times pd}, W^Q, W^K, W^V\in \mathbb{R}^{pd\times d}, W^O \in \mathbb{R}^{pd\times pd}$. And the dimensions of the two FC layers in the FFN are with dimensions $W_1 \in \mathbb{R}^{pd\times 4pd}, W_2 \in \mathbb{R}^{4pd \times pd}$. And we assume a sequence of input tokens with length $N$ is batched to process together. Since the computation for each attention head is computed independently, we only analyze the computation complexity of a single head attention, which is
$\mathcal{O}\big(\underbrace{d \cdot pd\cdot N}_{\text{proj. of $Q, K, V$}}+\underbrace{2N^2\cdot d}_{\text{attention layer}}+\underbrace{pd \cdot pd\cdot N}_{\text{proj. of  the output of attention}}\big)=\mathcal{O}\big((p+p^2)Nd^2+N^2d\big)=\mathcal{O}\big(Np^2d^2+N^2d\big)$.
Similarly, the computation complexity for the FFN layer is $\mathcal{O}\big(\underbrace{4\times p^2 d^2N}_{xW_1}+\underbrace{4\times p^2 d^2N}_{xW_1 W_2}\big)$. For the low-rank attention layer, the computation complexity becomes \\
$\mathcal{O}\big(\underbrace{(dr+rpd)\cdot N}_{\text{low-rank proj. }}+\underbrace{(pdr + rpd) \cdot N}_{\text{low-rank proj. of the output}}+2N^2\cdot d\big)=\mathcal{O}\big((p+1)drN+2Ndpr+2N^2d\big)=\mathcal{O}\big(pdrN + N^2d\big)$
and the computation complexity for FFN $\mathcal{O}\big(\underbrace{(p\cdot d\cdot r+4r\cdot h\cdot d)\cdot N}_{xW_1}+\underbrace{(p\cdot d\cdot r+4r\cdot p\cdot d)\cdot N}_{xW_1 W_2}\big)$. For the number of parameters, the vanilla multi-head attention layer contains $3pd^2\cdot p + p^2d^2=4p^2d^2$ parameters in total while the low-rank multi-head attention layer contains $3p(pdr+rd)+(pdr+rpd)=prd(3p+5)$. parameters in total. The vanilla FFN layer contains $4p^2d^2+4p^2d^2=8p^2d^2$ parameters in total while the low-rank FFN layer contains $(pdr+r4pd)+(4pdr+rpd)=10pdr$. parameters in total.

\section{Details on the dataset and models used for the experiment}
The details of the datasets used in the experiments are summarized in Table \ref{table:dataset-model}.
\begin{table*}[ht]
	\caption{The datasets used and their associated learning models.}
	\label{table:dataset-model}
	\begin{center}
	  \scriptsize{
		\begin{tabular}{ccccc}
		\toprule \textbf{Method}
		& CIFAR-10 &  ImageNet & WikiText-2 & WMT16' Gen-Eng 
		\bigstrut\\
		\midrule
		\# Data points & $60,000$ & $1,281,167$ & $29,000$ & $1,017,981$ \bigstrut\\
        Data Dimension & $32\times32\times3$ & $224\times224\times3$ & $1,500$ & $9,521$ \bigstrut\\
		Model & VGG-19-BN;ResNet-18 & ResNet-50;WideResNet-50-2 & 2 layer LSTM & Transformer ($p=8, N=6$)  \bigstrut\\
		Optimizer & \multicolumn{2}{c}{\textsc{SGD} } & \textsc{SGD} & \textsc{Adam} \bigstrut\\
		Hyper-params. & \multicolumn{2}{c}{Init lr: $0.01$ } & lr: $20$(decay with $0.25$ when val. loss not decreasing) & Init lr: 0.001 \bigstrut\\
		 \multicolumn{3}{c}{momentum: 0.9, $\ell_2$ weight decay: $10^{-4}$} & grad. norm clipping $0.25$ & $\beta s=(0.9, 0.98), \epsilon=10^{-8}$ \bigstrut\\		
		\bottomrule
		\end{tabular}}%
	\end{center}
\end{table*}

\section{Details on the hybrid networks in the experiments}
\paragraph{The hybrid VGG-19-BN architecture.} we generally found that using $K=10$ in the VGG-19-BN architecture leads to good test accuracy and moderate model compression ratio.
\begin{table}[H]
    \vspace{-4 mm}
	\caption{Detailed information of the hybrid VGG-19-BN architecture used in our experiments, all non-linear activation function in this architecture is ReLU after each convolution layer (omitted in the Table). The shapes for convolution layers follows $(c_{in}, c_{out}, k, k)$. There is a BatchNorm layer after each convolution layer with number of neurons the same as $c_{\text{out}}$ (also omitted in the Table).}
	\label{table:supp_vgg_architecture}
	\begin{center}
		 \scriptsize{
			\begin{tabular}{ccc}
				\toprule \textbf{Parameter}
				& Shape &  Layer hyper-parameter \bigstrut\\
				\midrule
				\textbf{layer1.conv1.weight} & $3 \times 64 \times 3 \times 3$ & stride:$1$;padding:$1$ \bigstrut\\
				\textbf{layer2.conv2.weight} & $64 \times 64 \times 3 \times 3$ & stride:$1$;padding:$1$  \bigstrut\\
				\textbf{pooling.max} & N/A & kernel size:$2$;stride:$2$  \bigstrut\\
				\textbf{layer3.conv3.weight} & $64\times 128 \times 3 \times 3$ & stride:$1$;padding:$1$ \bigstrut\\
				\textbf{layer4.conv4.weight} & $128\times 128 \times 3 \times 3$ & stride:$1$;padding:$1$ \bigstrut\\
                \textbf{pooling.max} & N/A & kernel size:$2$;stride:$2$  \bigstrut\\
				\textbf{layer5.conv5.weight} & $128 \times 256 \times 3 \times 3$ & stride:$1$;padding:$1$  \bigstrut\\
				\textbf{layer6.conv6.weight} & $256\times 256 \times 3 \times 3$ & stride:$1$;padding:$1$  \bigstrut\\
				\textbf{layer7.conv7.weight} & $256 \times 256 \times 3 \times 3$ & stride:$1$;padding:$1$  \bigstrut\\
				\textbf{layer8.conv8.weight} & $256 \times 256 \times 3 \times 3$ & stride:$1$;padding:$1$  \bigstrut\\
                \textbf{pooling.max} & N/A & kernel size:$2$;stride:$2$  \bigstrut\\
				\textbf{layer9.conv9.weight} & $256 \times 512 \times 3 \times 3$ & stride:$1$;padding:$1$  \bigstrut\\
				\textbf{layer10.conv10\_u.weight} & $512 \times 128 \times 3 \times 3$ & stride:$1$;padding:$1$  \bigstrut\\
				\textbf{layer10.conv10\_v.weight} & $128 \times 512 \times 1 \times 1$ & stride:$1$  \bigstrut\\
				\textbf{layer11.conv11\_u.weight} & $512 \times 128 \times 3 \times 3$ & stride:$1$;padding:$1$  \bigstrut\\
				\textbf{layer11.conv11\_v.weight} & $128 \times 512 \times 1 \times 1$ & stride:$1$  \bigstrut\\
				\textbf{layer12.conv12\_u.weight} & $512 \times 128 \times 3 \times 3$ & stride:$1$;padding:$1$  \bigstrut\\
				\textbf{layer12.conv12\_v.weight} & $128 \times 512 \times 1 \times 1$ & stride:$1$  \bigstrut\\
				\textbf{pooling.max} & N/A & kernel size:$2$;stride:$2$  \bigstrut\\
				\textbf{layer13.conv13\_u.weight} & $512 \times 128 \times 3 \times 3$ & stride:$1$;padding:$1$  \bigstrut\\
                \textbf{layer13.conv13\_v.weight} & $128 \times 512 \times 1 \times 1$ & stride:$1$  \bigstrut\\
				\textbf{layer14.conv14\_u.weight} & $512 \times 128 \times 3 \times 3$ & stride:$1$;padding:$1$  \bigstrut\\
				\textbf{layer14.conv14\_v.weight} & $128 \times 512 \times 1 \times 1$ & stride:$1$  \bigstrut\\
				\textbf{layer15.conv15\_u.weight} & $512 \times 128 \times 3 \times 3$ & stride:$1$;padding:$1$  \bigstrut\\
				\textbf{layer15.conv15\_v.weight} & $128 \times 512 \times 1 \times 1$ & stride:$1$  \bigstrut\\
				\textbf{layer16.conv16\_u.weight} & $512 \times 128 \times 3 \times 3$ & stride:$1$;padding:$1$  \bigstrut\\
				\textbf{layer16.conv16\_v.weight} & $128 \times 512 \times 1 \times 1$ & stride:$1$  \bigstrut\\
				\textbf{pooling.max} & N/A & kernel size:$2$;stride:$2$  \bigstrut\\
				\textbf{layer17.fc17.weight} & $512 \times 512$ & N/A  \bigstrut\\
				\textbf{layer17.fc17.bias} & $512$ & N/A  \bigstrut\\
				\textbf{layer18.fc18.weight} & $512 \times 512$ & N/A  \bigstrut\\
				\textbf{layer18.fc18.bias} & $512$ & N/A  \bigstrut\\
				\textbf{layer19.fc19.weight} & $512 \times 10$ & N/A  \bigstrut\\
				\textbf{layer19.fc19.bias} & $10$ & N/A  \bigstrut\\
				\bottomrule
			\end{tabular}}%
	\end{center}
\end{table}

\paragraph{The low-rank LSTM architecture.}
Note that we only use a 2-layer stacked LSTM as the model in the WikiText-2 next word prediction task. Our implementation is directly modified from the PyTorch original example \footnote{\url{https://github.com/pytorch/examples/tree/master/word_language_model}}. We used the tied version of LSTM, \ie enabling weight sharing for the encoder and decoder layers.
\begin{table}[H]
	\caption{Detailed information on the low-rank LSTM architecture in our experiment.}
	\label{table:architecture-lstm}
	\begin{center}
		 \scriptsize{
			\begin{tabular}{ccc}
				\toprule \textbf{Parameter}
				& Shape & Hyper-param. \bigstrut\\
				\midrule
				\textbf{encoder.weight} & $33278\times 1500$ & N/A \bigstrut\\
				\textbf{dropout} & N/A & $p=0.65$ \bigstrut\\
				\textbf{lstm0.weight.ii/f/g/o\_u} & $1500\times 375$ & N/A \bigstrut\\
				\textbf{lstm0.weight.ii/f/g/o\_v} & $375 \times 1500$ & N/A \bigstrut\\
				\textbf{lstm0.weight.hi/f/g/o\_u} & $1500\times 375$ & N/A \bigstrut\\
				\textbf{lstm0.weight.hi/f/g/o\_v} & $375 \times 1500$ & N/A \bigstrut\\
				\textbf{dropout} & N/A & $p=0.65$ \bigstrut\\
				\textbf{lstm1.weight.ii/f/g/o\_u} & $1500\times 375$ & N/A \bigstrut\\
				\textbf{lstm1.weight.ii/f/g/o\_v} & $375 \times 1500$ & N/A \bigstrut\\
				\textbf{lstm1.weight.hi/f/g/o\_u} & $1500\times 375$ & N/A \bigstrut\\
				\textbf{lstm1.weight.hi/f/g/o\_v} & $375 \times 1500$ & N/A \bigstrut\\
				\textbf{decoder.weight}(shared) & $1500\times 33278$ & N/A \bigstrut\\
				\bottomrule
			\end{tabular}}%
	\end{center}
\end{table}

\paragraph{The hybrid ResNet-18, ResNet-50, WideResNet-50-2 architectures.} For the CIFAR-10 dataset, we modified the original ResNet-50 architecture described in the original ResNet paper \citep{he2016deep}. The details about the modified ResNet-18 architecture for the CIFAR-10 dataset are shown in Table~\ref{tab:resnet18-cifar10-arch}. The network architecture is modified from the public code repository \footnote{\url{https://github.com/kuangliu/pytorch-cifar}}. For the first $2$ convolution block, \ie conv2\_x, we used stride at $1$ and padding at $1$ for all the convolution layers. For conv3\_x, conv4\_x, and conv5\_x we used the stride at $2$ and padding at $1$. We also note that there is a BatchNorm layer after each convolution layer with the number of elements equals the number of convolution filters. As shown in Table \ref{tab:resnet18-cifar10-arch}, our hybrid architecture starts from the $2$nd convolution block, \ie $K=4$. Our experimental study generally shows that this choice of hybrid ResNet-18 architecture leads to a good balance between the final model accuracy and the number of parameters. Moreover, we did not handle the downsample weights in the convolution blocks.
\newcommand{\blocka}[2]{\multirow{3}{*}{\(\left[\begin{array}{c}\text{3$\times$3, #1}\\[-.1em] \text{3$\times$3, #1} \end{array}\right]\)$\times$#2}
}
\newcommand{\blockb}[3]{\multirow{3}{*}{\(\left[\begin{array}{c}\text{1$\times$1, #2}\\[-.1em] \text{3$\times$3, #2}\\[-.1em] \text{1$\times$1, #1}\end{array}\right]\)$\times$#3}
}
\renewcommand\arraystretch{1.1}
\setlength{\tabcolsep}{3pt}
\begin{table}[H]
\begin{center}
\resizebox{1.0\linewidth}{!}{
\begin{tabular}{ccccccc}
\toprule
Layer Name & ResNet-18 & Rank Information  \\
\midrule
conv1 & \multicolumn{1}{c}{3$\times$3, 64, stride 1, padding 1} & full-rank\\
\midrule
\multirow{3}{*}{conv2\_x} & 
  \blocka{64}{2} & 1st block full-rank \\
  &  & 2nd block low-rank &  &  &  &\\
  &  & conv\_u $(64,16,3,3)$, conv\_v$(16, 64, 1, 1)$ &  &  &  &\\
\midrule
\multirow{3}{*}{conv3\_x}  & \blocka{128}{2} & low-rank
                              \\
  &  & conv\_u $(128, 32, 3, 3)$ &  &  &  & \\
  &  & conv\_v $(32, 128, 1, 1)$ &  &  &  & \\
\midrule
\multirow{3}{*}{conv4\_x} & \blocka{256}{2} & low-rank \\
  &  & conv\_u $(256, 64, 3, 3)$ &  &  & \\
  &  & conv\_v $(64, 256, 1, 1)$ &  &  & \\
\midrule
\multirow{3}{*}{conv5\_x}  & \blocka{512}{2} & low-rank \\
  &  & conv\_u $(512, 128, 3, 3)$ &  &  &  & \\
  &  & conv\_v $(128, 512, 1, 1)$ &  &  &  & \\
\midrule & \multicolumn{1}{c}{Avg Pool, 10-dim FC, SoftMax} \\
\bottomrule
\end{tabular}
}
\end{center}
\caption{The ResNet-18 architecture for the CIFAR-10 dataset used in the experiments.
}
\label{tab:resnet18-cifar10-arch}
\end{table}

For the ResNet-50 architecture, the detailed information is shown in the Table~\ref{tab:resnet50-imagenet-arch}. As we observed that the last three convolution blocks, \ie conv5\_x contains around $60\%$ of the total number of parameters in the entire network, thus we just put the last three convolution blocks as low-rank blocks and all other convolution blocks are full-rank blocks. Note that, different from the ResNet-18 architecture for the CIFAR-10 dataset described above. We also handle the downsample weight inside the ResNet-50 network, which only contains in the very first convolution block of conv5\_x. The original dimension of the downsample weight is with shape $(1024, 2048, 1, 1)$. Our factorization strategy leads to the shape of conv\_u: $(1024, 256, 1, 1)$ and conv\_v: $(256, 2048, 1, 1)$.
\begin{table}[H]
\begin{center}
\resizebox{1.0\linewidth}{!}{
\begin{tabular}{ccccccc}
\toprule
Layer Name & output size & ResNet-50 & Rank Information  \\
\midrule
conv1 & 112$\times$112 & \multicolumn{1}{c}{7$\times$7, 64, stride 2} & full-rank\\
\midrule
\multirow{4}{*}{conv2\_x} & \multirow{4}{*}{56$\times$56} & \multicolumn{1}{c}{3$\times$3 max pool, stride 2} \\\cline{3-7}
  &  & \blockb{256}{64}{3} \\
  &  &  & all blocks full-rank &  &  &\\
  &  &  &  &  &  &\\
\hline
\multirow{3}{*}{conv3\_x} &  \multirow{3}{*}{28$\times$28}   & \blockb{512}{128}{4}  &   &
                              \\
  &  &  & all blocks full-rank &  &  & \\
  &  &  &  &  &  & \\
\hline
\multirow{3}{*}{conv4\_x} & \multirow{3}{*}{14$\times$14}  & \blockb{1024}{256}{6}  \\
  &  &  & all blocks full-rank &  & \\
  &  &  &  &  & \\
\hline
\multirow{3}{*}{conv5\_x} & \multirow{3}{*}{7$\times$7} & \blockb{2048}{512}{3} & conv\_1\_u $(c_\text{in}, \frac{c_\text{in}}{4}, 1, 1)$; conv\_1\_v $(\frac{c_\text{in}}{4}, 512, 1, 1)$ \\
  &  &  & conv\_2\_u $(512, 128, 3, 3)$; conv\_2\_v $(128, 512, 1, 1)$  &  &  &  \\
  &  &  & conv\_3\_u $(512, 128, 1, 1)$; conv\_2\_v $(128, 2048, 1, 1)$ &  &  & \\
\hline
& 1$\times$1  & \multicolumn{1}{c}{Avg pool, 1000-dim FC, SoftMax} \\
\bottomrule
\end{tabular}
}
\end{center}
\caption{The ResNet-50 architecture for the ImageNet dataset used in the experiments.
}
\label{tab:resnet50-imagenet-arch}
\end{table}
For the WideResNet-50 architecture, the detailed architecture we used is shown in Table \ref{tab:wideresnet50-imagenet-arch}. Similar to what we observed for the ResNet-50 architecture, we just put the last three convolution blocks as low-rank blocks and all other convolution blocks are full-rank blocks. We also handle the downsample weight inside the WideResNet-50 network, which only contains the very first convolution block of conv5\_x. The original dimension of the downsample weight is with shape $(1024, 2048, 1, 1)$. Our factorization strategy leads to the shape of conv\_u: $(1024, 256, 1, 1)$ and conv\_v: $(256, 2048, 1, 1)$.
\begin{table}[H]
\begin{center}
\resizebox{1.0\linewidth}{!}{
\begin{tabular}{ccccccc}
\toprule
Layer Name & output size & WideResNet-50-2 & Rank Information  \\
\midrule
conv1 & 112$\times$112 & \multicolumn{1}{c}{7$\times$7, 64, stride 2} & full-rank\\
\midrule
\multirow{4}{*}{conv2\_x} & \multirow{4}{*}{56$\times$56} & \multicolumn{1}{c}{3$\times$3 max pool, stride 2} \\\cline{3-7}
  &  & \blockb{256}{128}{3} \\
  &  &  & all blocks full-rank &  &  &\\
  &  &  &  &  &  &\\
\hline
\multirow{3}{*}{conv3\_x} &  \multirow{3}{*}{28$\times$28}   & \blockb{512}{256}{4}  &   &
                              \\
  &  &  & all blocks full-rank &  &  & \\
  &  &  &  &  &  & \\
\hline
\multirow{3}{*}{conv4\_x} & \multirow{3}{*}{14$\times$14}  & \blockb{1024}{512}{6}  \\
  &  &  & all blocks full-rank &  & \\
  &  &  &  &  & \\
\hline
\multirow{3}{*}{conv5\_x} & \multirow{3}{*}{7$\times$7} & \blockb{2048}{1024}{3} & conv\_1\_u $(c_\text{in}, \frac{c_\text{in}}{4}, 1, 1)$; conv\_1\_v $(\frac{c_\text{in}}{4}, 1024, 1, 1)$ \\
  &  &  & conv\_2\_u $(1024, 256, 3, 3)$; conv\_2\_v $(256, 1024, 1, 1)$  &  &  &  \\
  &  &  & conv\_3\_u $(1024, 256, 1, 1)$; conv\_2\_v $(256, 2048, 1, 1)$ &  &  & \\
\hline
& 1$\times$1  & \multicolumn{1}{c}{Avg pool, 1000-dim FC, SoftMax} \\
\bottomrule
\end{tabular}
}
\end{center}
\caption{The WideResNet-50-2 architecture for the ImageNet dataset used in the experiments.
}
\label{tab:wideresnet50-imagenet-arch}
\end{table}

\paragraph{The hybrid Transformer architecture.} The Transformer architecture used in the experiment follows from the original Transformer paper \citep{vaswani2017attention}. Our implementation is modified from the public code repository \footnote{\url{https://github.com/jadore801120/attention-is-all-you-need-pytorch}}. We use the stack of $N=6$ encoder and decoder layers inside the Transformer architecture and number of head $p=8$. Since the encoder and decoder layers are identical across the entire architecture, we describe the detailed encoder and decoder architecture information in Table \ref{table:architecture-transformer-encoder} and Table \ref{table:architecture-transformer-decoder}. For the hybrid architecture used in the Transformer architecture, we put the very first encoder layer and first decoder layer as full-rank layers, and all other layers are low-rank layers. For low-rank encoder and decoder layers, we used the rank ratio at $\frac{1}{4}$, thus the shape of $U^Q, U^K, U^V, U^O \in \mathbb{R}^{512 \times 128}, V^{Q\top}, V^{K\top}, V^{V\top}, V^{O\top} \in \mathbb{R}^{128 \times 512}$. For $W_1$ in the $\text{FFN}(\cdot)$ layer, the $U_1\in \mathbb{R}^{512\times 128}, V_1^\top \in \mathbb{R}^{128\times 2048}$. For $W_2$ in the $\text{FFN}(\cdot)$ layer, the $U_2\in \mathbb{R}^{2048\times 128}, V_1^\top \in \mathbb{R}^{128\times 512}$. 

\begin{table}[H]
	\caption{Detailed information of the encoder layer in the Transformer architecture in our experiment}
	\label{table:architecture-transformer-encoder}
	\begin{center}
		 \scriptsize{
			\begin{tabular}{ccc}
				\toprule \textbf{Parameter}
				& Shape & Hyper-param. \bigstrut\\
				\midrule
				\textbf{embedding} & $9521\times 512$ & padding index: 1 \bigstrut\\
				\textbf{positional encoding} & N/A & N/A \bigstrut\\
				\textbf{dropout} & N/A & $p=0.1$ \bigstrut\\
				\textbf{encoder.self-attention.wq}($W^Q$) & $512\times 512$ & N/A \bigstrut\\
                \textbf{encoder.self-attention.wk}($W^K$) & $512\times 512$ & N/A \bigstrut\\
                \textbf{encoder.self-attention.wv}($W^V$) & $512\times 512$ & N/A \bigstrut\\
                \textbf{encoder.self-attention.wo}($W^O$) & $512\times 512$ & N/A \bigstrut\\
                \textbf{encoder.self-attention.dropout} & N/A & $p=0.1$ \bigstrut\\
                \textbf{encoder.self-attention.layernorm} & $512$ & $\epsilon=10^{-6}$
                \bigstrut\\
                \textbf{encoder.ffn.layer1} & $512\times 2048$ & N/A
                \bigstrut\\
                \textbf{encoder.ffn.layer2} & $2048\times 512$ & N/A
                \bigstrut\\
                \textbf{encoder.layernorm} & $512$ & $\epsilon=10^{-6}$
                \bigstrut\\
				\textbf{dropout} & N/A & $p=0.1$ \bigstrut\\
				\bottomrule
			\end{tabular}}%
	\end{center}
\end{table}

\begin{table}[H]
	\caption{Detailed information of the decoder layer in the Transformer architecture in our experiment}
	\label{table:architecture-transformer-decoder}
	\begin{center}
		 \scriptsize{
			\begin{tabular}{ccc}
				\toprule \textbf{Parameter}
				& Shape & Hyper-param. \bigstrut\\
				\midrule
				\textbf{embedding} & $9521\times 512$ & padding index: 1 \bigstrut\\
				\textbf{positional encoding} & N/A & N/A \bigstrut\\
				\textbf{dropout} & N/A & $p=0.1$ \bigstrut\\
				\textbf{decoder.self-attention.wq}($W^Q$) & $512\times 512$ & N/A \bigstrut\\
                \textbf{decoder.self-attention.wk}($W^K$) & $512\times 512$ & N/A \bigstrut\\
                \textbf{decoder.self-attention.wv}($W^V$) & $512\times 512$ & N/A \bigstrut\\
                \textbf{decoder.self-attention.wo}($W^O$) & $512\times 512$ & N/A \bigstrut\\
                \textbf{decoder.self-attention.dropout} & N/A & $p=0.1$ \bigstrut\\
                \textbf{decoder.self-attention.layernorm} & $512$ & $\epsilon=10^{-6}$
                \bigstrut\\
				\textbf{decoder.enc-attention.wq}($W^Q$) & $512\times 512$ & N/A \bigstrut\\
                \textbf{decoder.enc-attention.wk}($W^K$) & $512\times 512$ & N/A \bigstrut\\
                \textbf{decoder.enc-attention.wv}($W^V$) & $512\times 512$ & N/A \bigstrut\\
                \textbf{decoder.enc-attention.wo}($W^O$) & $512\times 512$ & N/A \bigstrut\\
                \textbf{decoder.enc-attention.dropout} & N/A & $p=0.1$ \bigstrut\\
                \textbf{decoder.enc-attention.layernorm} & $512$ & $\epsilon=10^{-6}$
                \bigstrut\\
                \textbf{decoder.ffn.layer1} & $512\times 2048$ & N/A
                \bigstrut\\
                \textbf{decoder.ffn.layer2} & $2048\times 512$ & N/A
                \bigstrut\\
                \textbf{encoder.layernorm} & $512$ & $\epsilon=10^{-6}$
                \bigstrut\\
				\textbf{dropout} & N/A & $p=0.1$ \bigstrut\\
				\bottomrule
			\end{tabular}}%
	\end{center}
\end{table}

\paragraph{The hybrid VGG-19-BN architecture used for the LTH comparison.} To compare \methodname{} with LTH, we use the open-source LTH implementation, \ie \url{https://github.com/facebookresearch/open_lth}. The VGG-19-BN model used in the open-source LTH repository is slightly different from the VGG-19-BN architecture described above. We thus use the VGG-19-BN architecture in the LTH code and deploy \methodname{} on top of it for fairer comparison. Detailed information about the hybrid VGG-19-BN architecture we used in \methodname{} for the comparison with LTH is shown in Table~\ref{table:supp_vgg_architecture_lth}.
\begin{table}[H]
    \vspace{-4 mm}
	\caption{Detailed information of the hybrid VGG-19-BN architecture used in our LTH comparison experiments, all non-linear activation function in this architecture is ReLU after each convolution layer (omitted in the Table). The shapes for convolution layers follows $(c_{in}, c_{out}, k, k)$. There is a BatchNorm layer after each convolution layer with number of neurons the same as $c_{\text{out}}$ (also omitted in the Table).}
	\label{table:supp_vgg_architecture_lth}
	\begin{center}
		 \scriptsize{
			\begin{tabular}{ccc}
				\toprule \textbf{Parameter}
				& Shape &  Layer hyper-parameter \bigstrut\\
				\midrule
				\textbf{layer1.conv1.weight} & $3 \times 64 \times 3 \times 3$ & stride:$1$;padding:$1$ \bigstrut\\
				\textbf{layer2.conv2.weight} & $64 \times 64 \times 3 \times 3$ & stride:$1$;padding:$1$  \bigstrut\\
				\textbf{pooling.max} & N/A & kernel size:$2$;stride:$2$  \bigstrut\\
				\textbf{layer3.conv3.weight} & $64\times 128 \times 3 \times 3$ & stride:$1$;padding:$1$ \bigstrut\\
				\textbf{layer4.conv4.weight} & $128\times 128 \times 3 \times 3$ & stride:$1$;padding:$1$ \bigstrut\\
                \textbf{pooling.max} & N/A & kernel size:$2$;stride:$2$  \bigstrut\\
				\textbf{layer5.conv5.weight} & $128 \times 256 \times 3 \times 3$ & stride:$1$;padding:$1$  \bigstrut\\
				\textbf{layer6.conv6.weight} & $256\times 256 \times 3 \times 3$ & stride:$1$;padding:$1$  \bigstrut\\
				\textbf{layer7.conv7.weight} & $256 \times 256 \times 3 \times 3$ & stride:$1$;padding:$1$  \bigstrut\\
				\textbf{layer8.conv8.weight} & $256 \times 256 \times 3 \times 3$ & stride:$1$;padding:$1$  \bigstrut\\
                \textbf{pooling.max} & N/A & kernel size:$2$;stride:$2$  \bigstrut\\
				\textbf{layer9.conv9.weight} & $256 \times 512 \times 3 \times 3$ & stride:$1$;padding:$1$  \bigstrut\\
				\textbf{layer10.conv10\_u.weight} & $512 \times 128 \times 3 \times 3$ & stride:$1$;padding:$1$  \bigstrut\\
				\textbf{layer10.conv10\_v.weight} & $128 \times 512 \times 1 \times 1$ & stride:$1$  \bigstrut\\
				\textbf{layer11.conv11\_u.weight} & $512 \times 128 \times 3 \times 3$ & stride:$1$;padding:$1$  \bigstrut\\
				\textbf{layer11.conv11\_v.weight} & $128 \times 512 \times 1 \times 1$ & stride:$1$  \bigstrut\\
				\textbf{layer12.conv12\_u.weight} & $512 \times 128 \times 3 \times 3$ & stride:$1$;padding:$1$  \bigstrut\\
				\textbf{layer12.conv12\_v.weight} & $128 \times 512 \times 1 \times 1$ & stride:$1$  \bigstrut\\
				\textbf{pooling.max} & N/A & kernel size:$2$;stride:$2$  \bigstrut\\
				\textbf{layer13.conv13\_u.weight} & $512 \times 128 \times 3 \times 3$ & stride:$1$;padding:$1$  \bigstrut\\
                \textbf{layer13.conv13\_v.weight} & $128 \times 512 \times 1 \times 1$ & stride:$1$  \bigstrut\\
				\textbf{layer14.conv14\_u.weight} & $512 \times 128 \times 3 \times 3$ & stride:$1$;padding:$1$  \bigstrut\\
				\textbf{layer14.conv14\_v.weight} & $128 \times 512 \times 1 \times 1$ & stride:$1$  \bigstrut\\
				\textbf{layer15.conv15\_u.weight} & $512 \times 128 \times 3 \times 3$ & stride:$1$;padding:$1$  \bigstrut\\
				\textbf{layer15.conv15\_v.weight} & $128 \times 512 \times 1 \times 1$ & stride:$1$  \bigstrut\\
				\textbf{layer16.conv16\_u.weight} & $512 \times 128 \times 3 \times 3$ & stride:$1$;padding:$1$  \bigstrut\\
				\textbf{layer16.conv16\_v.weight} & $128 \times 512 \times 1 \times 1$ & stride:$1$  \bigstrut\\
				\textbf{pooling.max} & N/A & kernel size:$2$;stride:$2$  \bigstrut\\
				\textbf{layer17.fc17.weight} & $512 \times 10$ & N/A  \bigstrut\\
				\textbf{layer17.fc17.bias} & $10$ & N/A  \bigstrut\\
				\bottomrule
			\end{tabular}}%
	\end{center}
\end{table}

\section{The compatibility of \methodname{} with other gradient compression methods}
\begin{figure}[ht]
    \vspace{-2 mm}
    \centering
    \subfigure[Breakdown per-epoch time]{\includegraphics[width=0.4\textwidth]{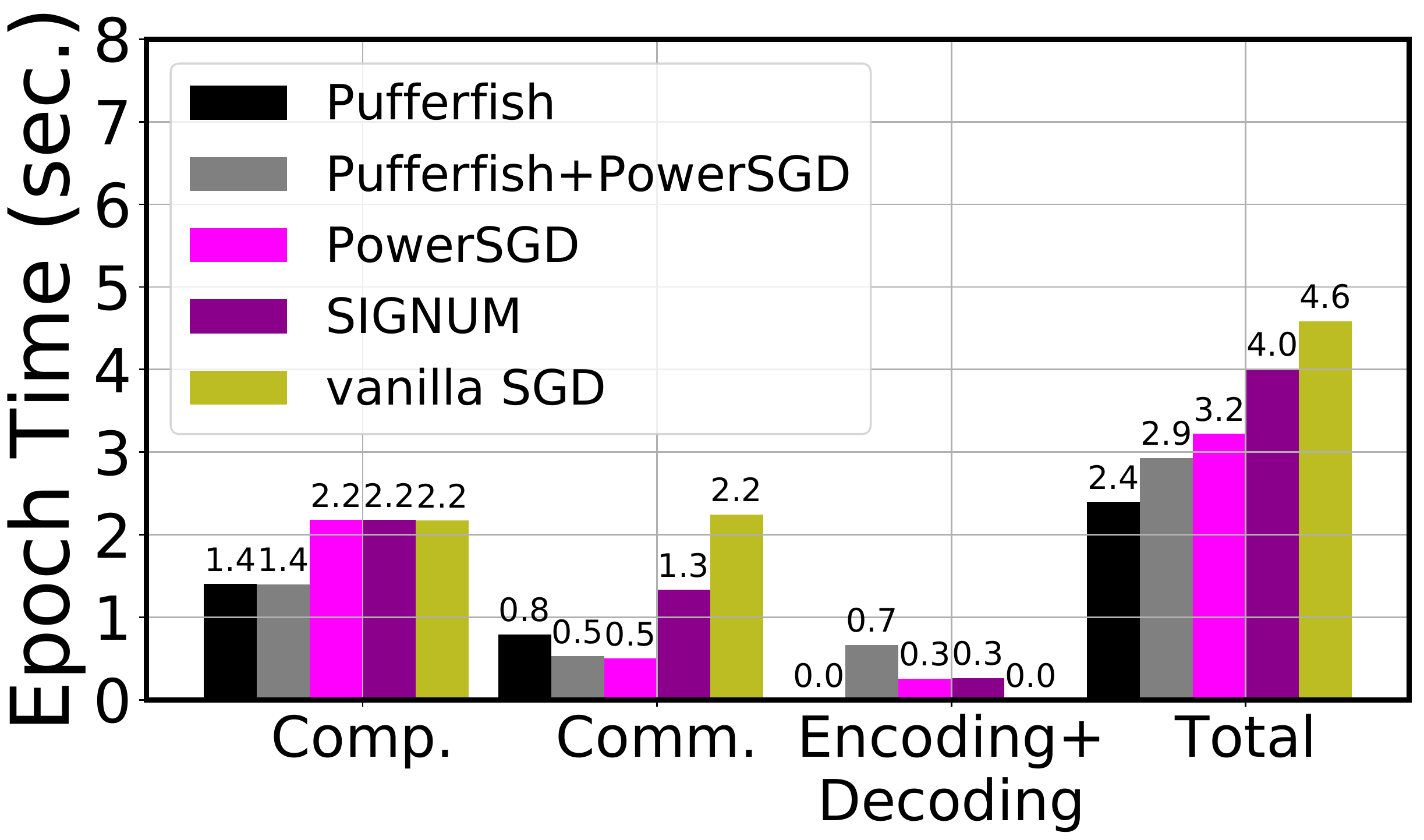}}\\
    \subfigure[Convergence]{\includegraphics[width=0.4\textwidth]{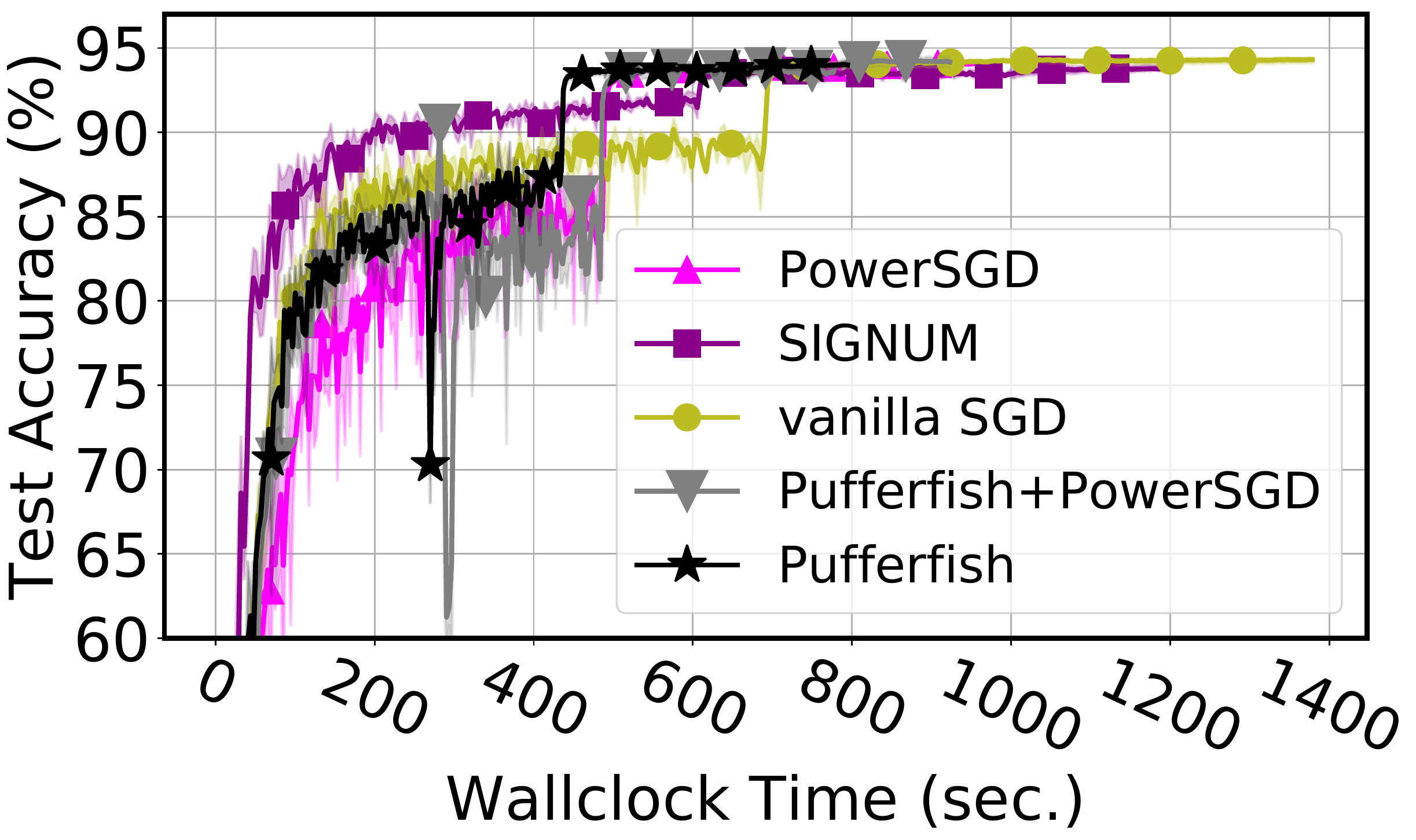}}
    \vspace{-4 mm}
    \caption{(a) Per-epoch breakdown runtime analysis and (b) convergence performance of \methodname{}, ``\methodname{}+\textsc{PowerSGD} (rank $4$)", \textsc{PowerSGD} (rank $2$), \textsc{signum}, and vanilla SGD over ResNet-18 trained on the CIFAR-10 dataset.
    }
    \label{fig:compat-pufferfish}
    \vspace{-4 mm}
    \end{figure}

As \methodname{} is a training time parameter reduction method, the gradient of the factorized networks can be compressed further with any gradient compression methods. As \textsc{PowerSGD} is the state-of-the-art gradient compression method and is compatible with \texttt{allreduce}, we consider another baseline, \ie ``\methodname{}+\textsc{PowerSGD}", we conduct an experimental study over this baseline on ResNet-18 trained on CIFAR-10 (results shown in Figure~\ref{fig:compat-pufferfish}). The experiment is running over $8$ \texttt{p3.2xlarge} EC2 nodes with batch size at $256$ per node ($2048$ in total). The experimental results indicate that combining \methodname{} with \textsc{PowerSGD} can effectively reduce the gradient size of \methodname{}, making \methodname{} enjoys high computation efficiency and the communication efficiency as high as \textsc{PowerSGD}. However, as \textsc{PowerSGD} conducts layer-wise gradient encoding and decoding on both $U_l$ and $V_l$ layers, the gradient encoding and decoding cost in the ``\methodname{}+\textsc{PowerSGD}" baseline is higher compared to \textsc{PowerSGD}. We observe that a slightly higher rank is desired when combining \methodname{} with \textsc{PowerSGD} since both model weights and gradients are approximated in this case. In the experimental results shown in Figure~\ref{fig:compat-pufferfish}, we use \textsc{PowerSGD} with rank $4$ when combining with \methodname{} for both the vanilla warm-up training epochs and the consecutive low-rank training epochs. Moreover, we also found that under the large-batch setting, it is always helpful to re-warmup the learning rate for the ``\methodname{}+\textsc{PowerSGD}" baseline, \ie in the first $5$ epochs, we warm-up the learning rate linearly from $0.1$ to $1.6$, then at the $80$-th epoch where we switch from the vanilla warm-up training to low-rank training, we repeat the learning rate warm-up again with $5$ epochs (from $0.1$ to $1.6$). Our experimental results suggest that \methodname{} can be combined with the gradient compression methods to attain better communication efficiency, but it is desirable to combine \methodname{} with the gradient compression methods that can be deployed on the fattened gradients, \eg Top-$k$.

\section{Discussion on the communication efficiency of \methodname{}}
It is natural to ask the question that ``\textit{Why are the previously proposed light weight gradient compression methods slow in practice, \eg the ones proposed in~\cite{suresh2016distributed}}?" We agree that there are lots of gradient compression methods, which are computationally cheap. However, other important factors can affect the gradient compression efficiency in practice (taking the gradient compression method in~\cite{suresh2016distributed} as an example): 

(i) After the binary sign rounding, extra encoding and decoding steps e.g. binary encoding are required to aggregate the quantized bits to bytes for attaining real communication speedup. That is optimizing the data structures to support low-communication for quantized gradients is necessary for any benefit to the surface, and also quite non trivial. 
(ii) For most gradient compression schemes, the encoded gradients are not compatible with all-reduce. Thus, all-gather has to be used instead. Unfortunately, in terms of comm. costs all-gather suffers a performance gap that increases with the number of nodes.
(iii) In all-reduce, each worker receives a pre-aggregated gradient, making the cost of decompression independent to the number of workers. In all-gather, a worker receives the number of workers compressed gradients that need to be individually decompressed and aggregated. The time for decompression with all-gather therefore scales linearly with the number of workers. 

\begin{figure}[ht]
    \vspace{-2 mm}
    \centering
    \includegraphics[width=0.4\textwidth]{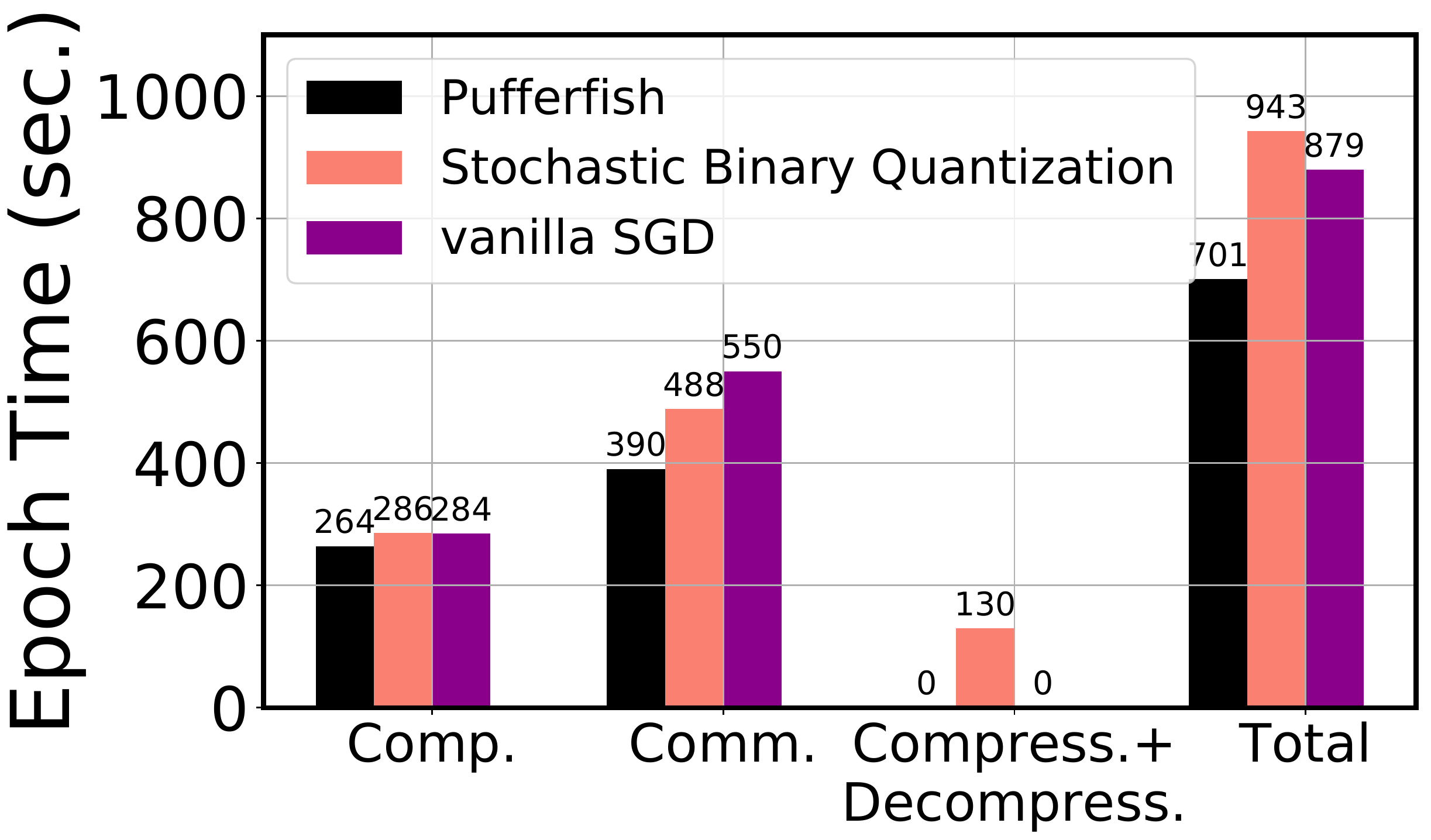}
    \vspace{-2 mm}
    \caption{Breakdown per-epoch runtime comparison between \methodname{}, vanilla SGD, and stochastic binary quantization.
    }
    \label{fig:sbq-comparisons}
    \vspace{-4 mm}
    \end{figure}

In fact we did run a test for the ``\textit{Stochastic binary quantization}" method in~\cite{suresh2016distributed} on ResNet-50+ImageNet over 16 EC2 \texttt{p3.2xlarge} nodes (per node batch size 32) as it is the computationally cheapest methods proposed in the paper. Though it is showed that conducting random rotation over the gradients can improve the compression error, we only care about the computational and communication efficiencies of the method in this particular experiment. Per epoch runtime results are shown in Figure~\ref{fig:sbq-comparisons}.

Note that in the ``compress.+decompress." stage, stochastic binary quantization takes $12.1\pm 0.6$ seconds for gradient compression and $118.4\pm 0.1$ for gradient decompression. We observe that although the stochastic binary quantization is efficient in the compression stage, its gradient decompression cost is expensive. Moreover, all-gather is less efficient compared to all-reduce at the scale of $16$ nodes.

\section{The effectiveness of using SVD to find the low-rank factorization} In the vanilla warm-up training strategy proposed in \methodname{}, we decompose the network weights using SVD to find the initialization weights for the hybrid network. Though SVD is a computationally expensive method, \methodname{} only requires to conduct the factorization over the network weights once during the entire training process. We explicitly test the overhead incurred by conducting SVD over the model weights here. All the runtimes are measured over the \texttt{p3.2xlarge} instance of Amazon EC2 (equipped with Tesla V100 GPU). The results are shown in Figure \ref{table:svd-efficiency}. From the results, it can be observed that the run time on using SVD to factorize the partially trained vanilla full-rank network is quite fast, \eg on average it only costs $2.2972$ seconds over the ResNet-50 trained over the ImageNet dataset, which only takes $0.17\%$ of the per epoch training time.
\begin{table}[ht]
	\caption{The time costs on conducting SVD over the partially trained vanilla full-rank network to find the initialization model for the hybrid network. The run time results are averaged from $5$ independent trials.}
	\label{table:svd-efficiency}
	\begin{center}
	  \scriptsize{
		\begin{tabular}{cc}
		\toprule \textbf{Method}
		& Time Cost (in sec.)  
		\bigstrut\\
		\midrule
		ResNet-50 on ImageNet & $2.2972\pm0.0519$ \bigstrut\\
		WideResNet-50-2 on ImageNet & $4.8700\pm0.0859$ \bigstrut\\
		VGG-19-BN on CIFAR-10 & $1.5198\pm0.0113$ \bigstrut\\
		ResNet-18 on CIFAR-10 & $1.3244\pm0.0201$ \bigstrut\\
		LSTM on WikeText-2 & $6.5791\pm0.0445$ \bigstrut\\
        Transformer on WMT16 & $5.4104\pm0.0532$ \bigstrut\\
		\bottomrule
		\end{tabular}}%
	\end{center}
\end{table}

\section{Details of data preprocessing}
\paragraph{The CIFAR-10 dataset.} In preprocessing the images in CIFAR-10 dataset, we follow the standard data augmentation and normalization process. For data augmentation, random cropping and horizontal random flipping are used. Each color channels are normalized with mean and standard deviation by $\mu_r = 0.491, \mu_g = 0.482, \mu_b =  0.447$, $\sigma_r = 0.247, \sigma_g = 0.244, \sigma_b = 0.262$. Each channel pixel is normalized by subtracting the mean value in this color channel and then divided by the standard deviation of this color channel.

\paragraph{The ImageNet dataset.} For ImageNet, we follow the data augmentation process of \citep{goyal2017accurate}, \textit{i.e.}, we use scale and aspect ratio data augmentation. The network input image is a $224\times 224$ pixels, randomly cropped from an augmented image or its horizontal flip. The input image is normalized in the same way as we normalize the CIFAR-10 images using the following  means and standard deviations: $\mu_r = 0.485, \mu_g = 0.456, \mu_b =  0.406$; $\sigma_r = 0.229, \sigma_g = 0.224, \sigma_b = 0.225$.

\section{Detailed hyper-parameters used in our experiments}
\paragraph{ResNet-50 and WideResNet-50-2 over the ImageNet dataset.} For ResNet-50 and WideResNet-50-2 models, we follow the model training hyper-parameters reported in \citep{goyal2017accurate}. We train the model using the optimizer SGD with momentum value at $0.9$ with batch size at $256$. We also conduct $\ell_2$ regularization over the model weights instead of the BatchNorm layers with the regularization coefficient $10^{-4}$. The entire training process takes $90$ epochs. For both of the ResNet-50 and WideResNet-50-2 models, we start from the learning rate at $0.1$ and decay the learning rate by a factor of $0.1$ at the $30$-th, $60$-th, and the $80$-th epochs. For the vanilla warm-up training, we use warm-up epoch $E=10$. Note that at the $10$-th epoch we switch from the vanilla ResNet-50/WideResNet-50-2 models to the hybrid architecture, but we still use the same learning rate, \ie $0.1$ until the $30$-th epoch. Additional to the previously proposed work, we adopt the \textit{label smoothing} technique with probability $0.1$. The model initialization method follows directly from the implementation of PyTorch example \footnote{\url{https://github.com/pytorch/examples/tree/master/imagenet}}.

\paragraph{ResNet-18 and VGG-19-BN over the CIFAR-10 dataset.} For ResNet-18 and VGG-19-BN models. We train the model using the optimizer SGD with momentum with momentum value at $0.9$ with batch size at $128$. The entire training takes $300$ epochs. We also conduct $\ell_2$ regularization over the model weights with the regularization coefficient $10^{-4}$. For both of the ResNet-18 and VGG-19-BN models, we start from the learning rate at $0.1$ and decay the learning rate by a factor of $0.1$ at the $150$-th, $250$-th epochs. For the vanilla warm-up training, we use warm-up epoch $E=80$. Note that at the $80$-th epoch we switch from the vanilla ResNet-18/VGG-19-BN models to the hybrid architecture, but we still use the same learning rate, \ie $0.1$ until the $150$-th epoch.

\paragraph{LSTM over the WikiText-2 dataset.} For the LSTM model. We conduct training using the vanilla SGD optimizer with batch size at $20$. We also conduct gradient norm clipping with norm bound at $0.25$. The entire training takes $40$ epochs. We start from the learning rate at $20$ and decay the learning rate by a factor of $0.25$ if the validation loss is not decreasing. For the vanilla warm-up training, we use warm-up epoch $E=10$. Note that at the $10$-th epoch we switch from the vanilla LSTM model to the hybrid architecture, we also decay the learning rate by a factor of $0.5$. We also tie the word embedding and SoftMax weights \citep{press2016using}.

\paragraph{The Transformer over the WMT16 dataset.} For the Transformer model. We conduct training using the Adam optimizer with initial learning rate at $0.001$, $\beta s=(0.9, 0.98), \epsilon=10^{-8}$ batch size at $256$. We also conduct gradient norm clipping with norm bound at $0.25$. The entire training takes $400$ epochs. For the vanilla warm-up training, we use warm-up epoch $E=10$. We enable label smoothing, weight sharing for the source and target word embedding, and weight sharing between target word embedding and the last dense layer.

\section{Detailed information on the runtime mini-benchmark}
In the experiment section, we discussed that in the reproducibility optimized setting, factorized networks achieve promising runtime speedup over the vanilla networks. However, sometimes users prefer faster runtime to reproducibility where the speed optimized setting is used (with \texttt{cudnn.benckmark} enabled and \texttt{cudnn.deterministic} disabled). We also study the runtime of the factorized network under the speed optimized setting. The results are shown in Table~\ref{table:mini-benchmark-benchmark}, from which we observe that the speedup of the factorized network is less promising compared to the reproducibility optimized setting especially for the VGG-19-BN network. However, \methodname{} ResNet-18 still achieves  $1.16\times$ per-epoch speedup. We leave exploring the optimal model training speed of the factorized networks as the future work.
\begin{table}[ht]
	\caption{The runtime mini-benckmark results of \methodname{} and vanilla VGG-19-BN and ResNet-18 networks training on the CIFAR-10 dataset, results averaged over $10$ epochs. Experiment running on a single V100 GPU with batch size at $128$; Over the optimized cuDNN implementation with \texttt{cudnn.benckmark} enabled and \texttt{cudnn.deterministic} disabled; Speedup calcuated based on the averaged per-epoch time.}
	\label{table:mini-benchmark-benchmark}
	\begin{center}
      \scriptsize{
		\begin{tabular}{cccc}
		\toprule \textbf{Model Archs.}
		&  Epoch Time (sec.) & Speedup & MACs (G) 
		\bigstrut\\
		\midrule
		Vanilla VGG-19 & $8.27 \pm 0.07$ & $-$ & $0.4$ \bigstrut\\
		\methodname{} VGG-19  & ${\bf 8.16}\pm 0.12$ & $\bf{1.01\times}$& $\bf{0.29}$ \bigstrut\\
		Vanilla ResNet-18  & $11.15\pm 0.01$ & $-$ & $0.56$ \bigstrut\\
		\methodname{} ResNet-18 & ${\bf9.61} \pm 0.08$ & $\bf{1.16\times}$ & $\bf{0.22}$ \bigstrut\\
		\bottomrule
		\end{tabular}}%
	\end{center}
\end{table}

\section{Time cost measurement on Amazon EC2}
We use the \texttt{p3.2xlarge} instances for the distributed experiments, the bandwidth of the instance is ``Up to $10$ Gbps" as stated on the Amazon EC2 website, \ie \url{https://aws.amazon.com/ec2/instance-types/p3/}. For some tasks (especially for the ResNet-50 and WideResNet-50-2), we observe that the bandwidth of the \texttt{p3.2xlarge} instance decays sharply in the middle of the experiment. The time costs for ResNet-50 trained on the ImageNet dataset under our prototype \texttt{allreduce} distributed implementation are collected when there is no bandwidth decay, \eg under $10$ Gbps. For the DDP time cost results, we run the experiments till the per-epoch time costs become stable, then measure the per-epoch time. For ResNet-18 trained on the CIFAR-10 dataset experiments under our prototype \texttt{allreduce} distributed implementation, we do not observe significant bandwidth decay for the \texttt{p3.2xlarge} instances. All distributed experiments are conducted under the \texttt{us-west-2c} availability zone of EC2.

\section{Additional experimental results}
\paragraph{The ablation study on the accuracy mitigation strategy over CIFAR-10 and ImageNet.} The ablation study results are shown in Table ~\ref{table:ablation-resnet50-imagenet} for ResNet-50 trained on ImageNet and Table~\ref{table:ablation-vgg19-cifar10} for VGG-19-BN trained over CIFAR-10. For the vanilla low-rank ResNet-50 trained on ImageNet, we do not deploy the label smoothing and the extra learning rate decay (with a factor $0.1$) at the $80$-th epoch.

\begin{table}[ht]
	\caption{The effect of vanilla warm-up training and hybrid network architectures of \methodname{} of the low-rank ResNet-50 trained over the ImageNet dataset}
	\label{table:ablation-resnet50-imagenet}
	\begin{center}
		 \scriptsize{
		\begin{tabular}{ccc}
		\toprule \textbf{Model architectures}
		 & Test Acc. Top1 & Test Acc. Top5 
		\bigstrut\\
		\midrule
		Low-rank ResNet-50 & $71.03\%$ & $90.26\%$ \bigstrut\\
		Hybrid ResNet-50 (wo. vanilla warm-up) & $75.85\%$ & $92.96\%$ \bigstrut\\
		Hybrid ResNet-50 (w. vanilla warm-up)  & $\bf{76.43}\%$ & $\bf{93.10}\%$ \bigstrut\\
		\bottomrule
		\end{tabular}%
		}
	\vspace{-6mm}
	\end{center}
\end{table}
\vspace{-2 mm}

\begin{table}[ht]
	\caption{The effect of vanilla warm-up training and hybrid network architectures of \methodname{} of the low rank VGG-19-BN trained over the CIFAR-10 dataset. Results are averaged across $3$ independent trials with different random seeds.}
	\label{table:ablation-vgg19-cifar10}
	\begin{center}
		 \scriptsize{
		\begin{tabular}{ccc}
		\toprule \textbf{Model architectures}
		 & Test Loss & Test Accuracy 
		\bigstrut\\
		\midrule
		Low-rank VGG-19-BN & $0.355\pm 0.012$ & $93.34\pm 0.08\%$ \bigstrut\\
		Hybrid VGG-19-BN (wo. vanilla warm-up) & $0.407\pm 0.008$ & $93.53\pm 0.13\%$ \bigstrut\\
		Hybrid VGG-19-BN (w. vanilla warm-up)  & $0.375\pm 0.019$ & ${\bf93.89} \pm 0.14\%$ \bigstrut\\
		\bottomrule
		\end{tabular}%
		}
	\end{center}
\end{table}

\end{document}